\documentclass{article}

\usepackage{microtype}
\usepackage{graphicx}
\usepackage{subfigure}
\usepackage{booktabs} 
\RequirePackage{subcaption}
\usepackage{rotating}
\usepackage{multirow}
\usepackage{array}
\usepackage{listings}
\usepackage{xcolor}
\usepackage{xspace}
\newcommand{\daoyuan}[1]{}

\usepackage{hyperref}

\usepackage[accepted]{icml2024}

\usepackage{amsmath}
\usepackage{amssymb}
\usepackage{mathtools}
\usepackage{amsthm}

\usepackage[capitalize,noabbrev]{cleveref}

\theoremstyle{plain}

\theoremstyle{definition}

\theoremstyle{remark}

\usepackage[textsize=tiny]{todonotes}

\newcommand{\ours}{\texttt{ArtAug}\xspace}

\icmltitlerunning{Submission and Formatting Instructions for ICML 2024}

\begin{document}

\twocolumn[
\icmltitle{\ours: Enhancing Text-to-Image Generation through \\
Synthesis-Understanding Interaction}




\begin{icmlauthorlist}
\icmlauthor{Zhongjie Duan}{ecnu}
\icmlauthor{Qianyi Zhao}{ecnu}
\icmlauthor{Cen Chen}{ecnu}
\icmlauthor{Daoyuan Chen}{ali}
\icmlauthor{Wenmeng Zhou}{ali}
\icmlauthor{Yaliang Li}{ali}
\icmlauthor{Yingda Chen}{ali}
\end{icmlauthorlist}

\icmlaffiliation{ecnu}{East China Normal University, Shanghai, China}
\icmlaffiliation{ali}{Alibaba Group, Hangzhou, China}

\icmlcorrespondingauthor{Cen Chen}{cenchen@dase.ecnu.edu.cn}

\icmlkeywords{Machine Learning, ICML}

\vskip 0.3in
]



\printAffiliationsAndNotice{}  

\begin{figure*}
  \centering
    \renewcommand{\arraystretch}{1.5}
    \setlength{\tabcolsep}{3pt}
    \begin{tabular}{m{0.1cm}<{\centering} m{0.233\textwidth}<{\centering} m{0.233\textwidth}<{\centering} m{0.233\textwidth}<{\centering} m{0.233\textwidth}<{\centering}}
        \rotatebox{90}{w/o \ours} & 
        \includegraphics[width=0.233\textwidth]{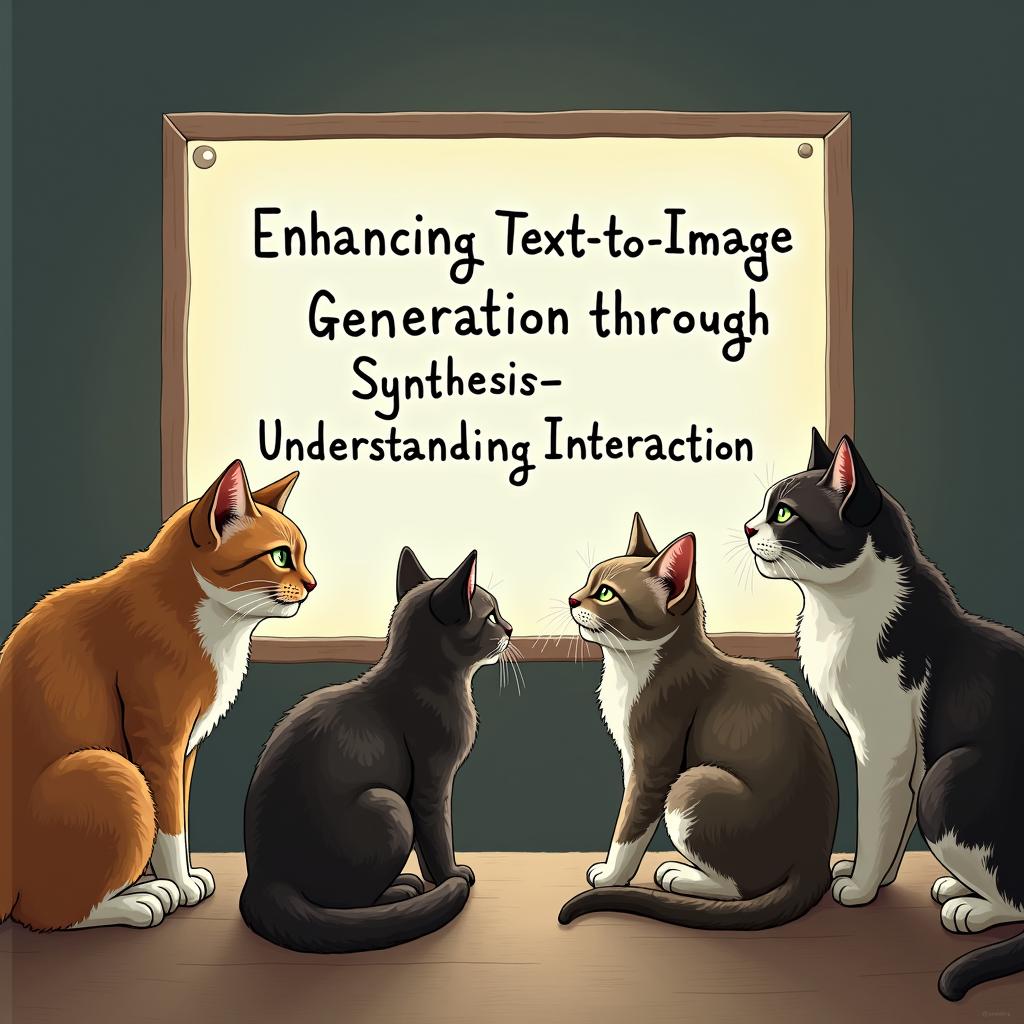} & 
        \includegraphics[width=0.233\textwidth]{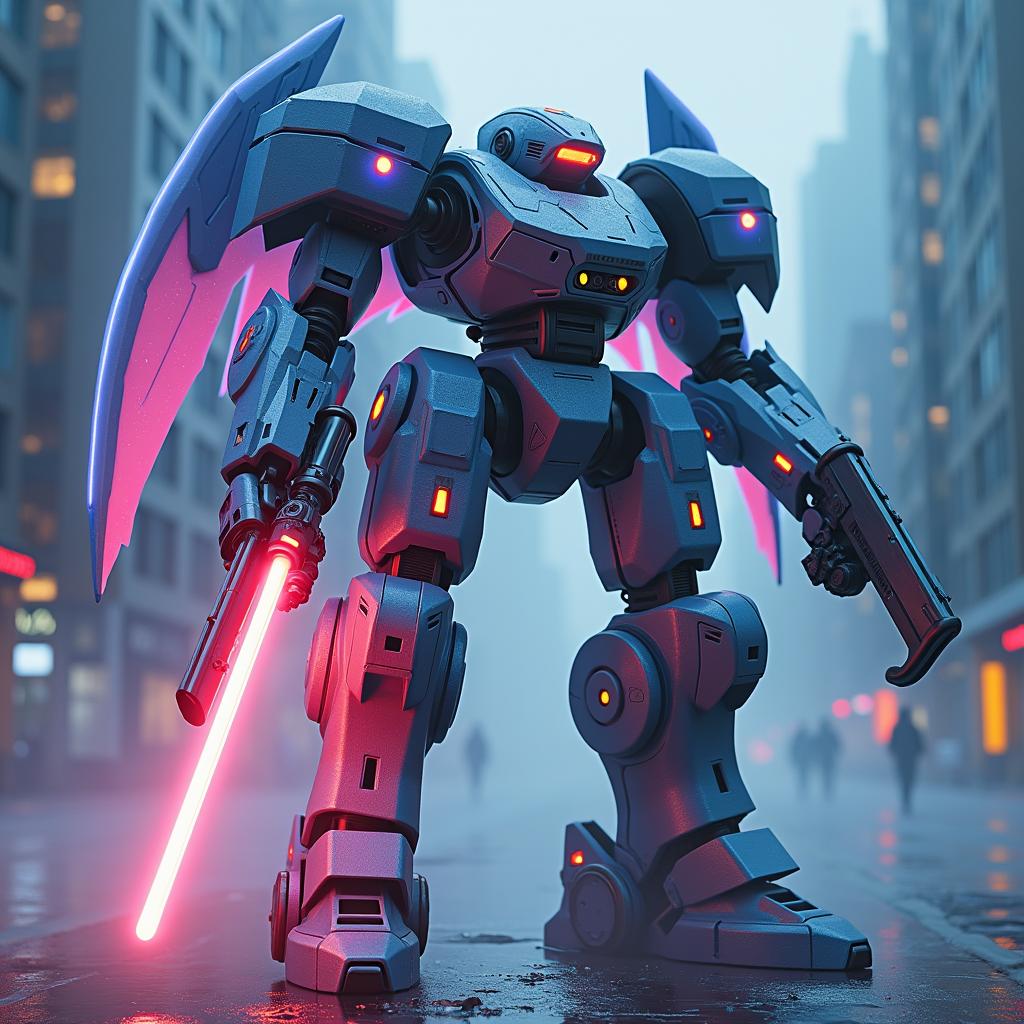} &
        \includegraphics[width=0.233\textwidth]{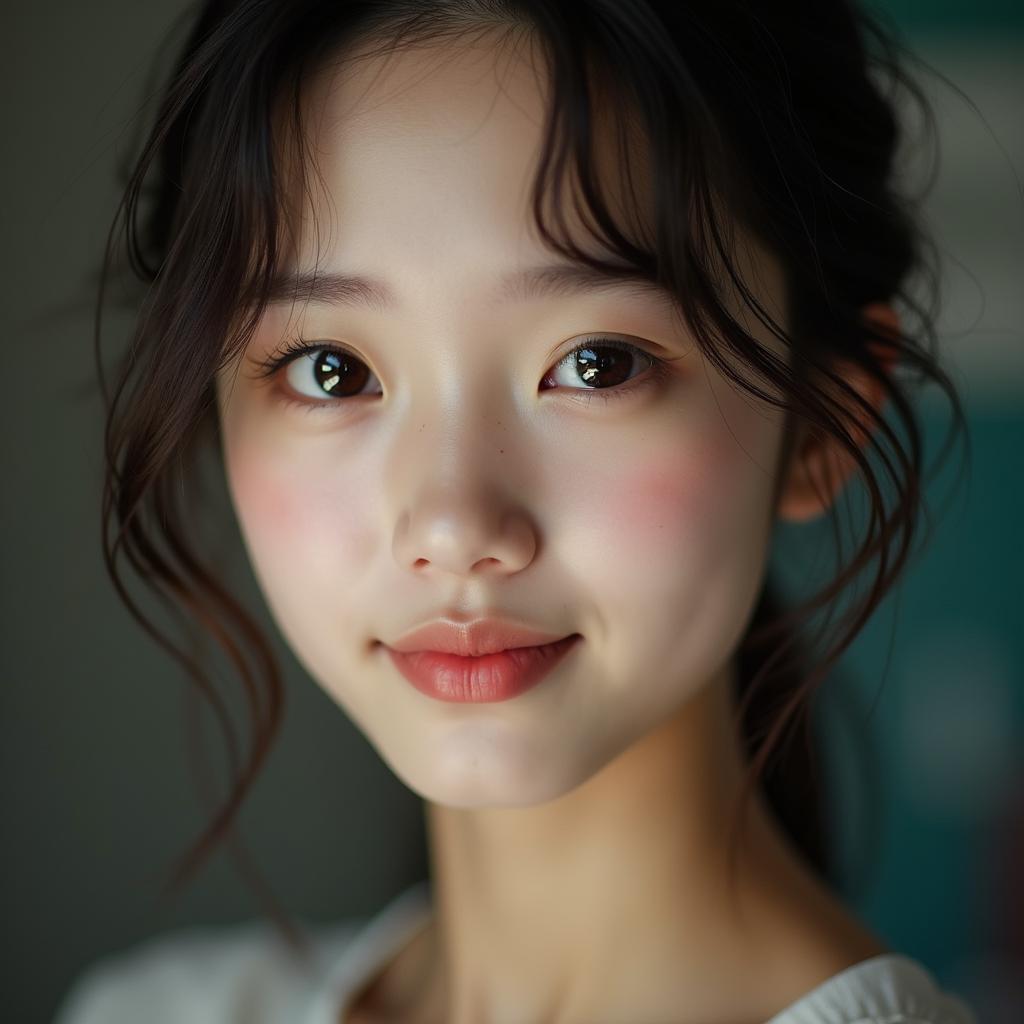} & 
        \includegraphics[width=0.233\textwidth]{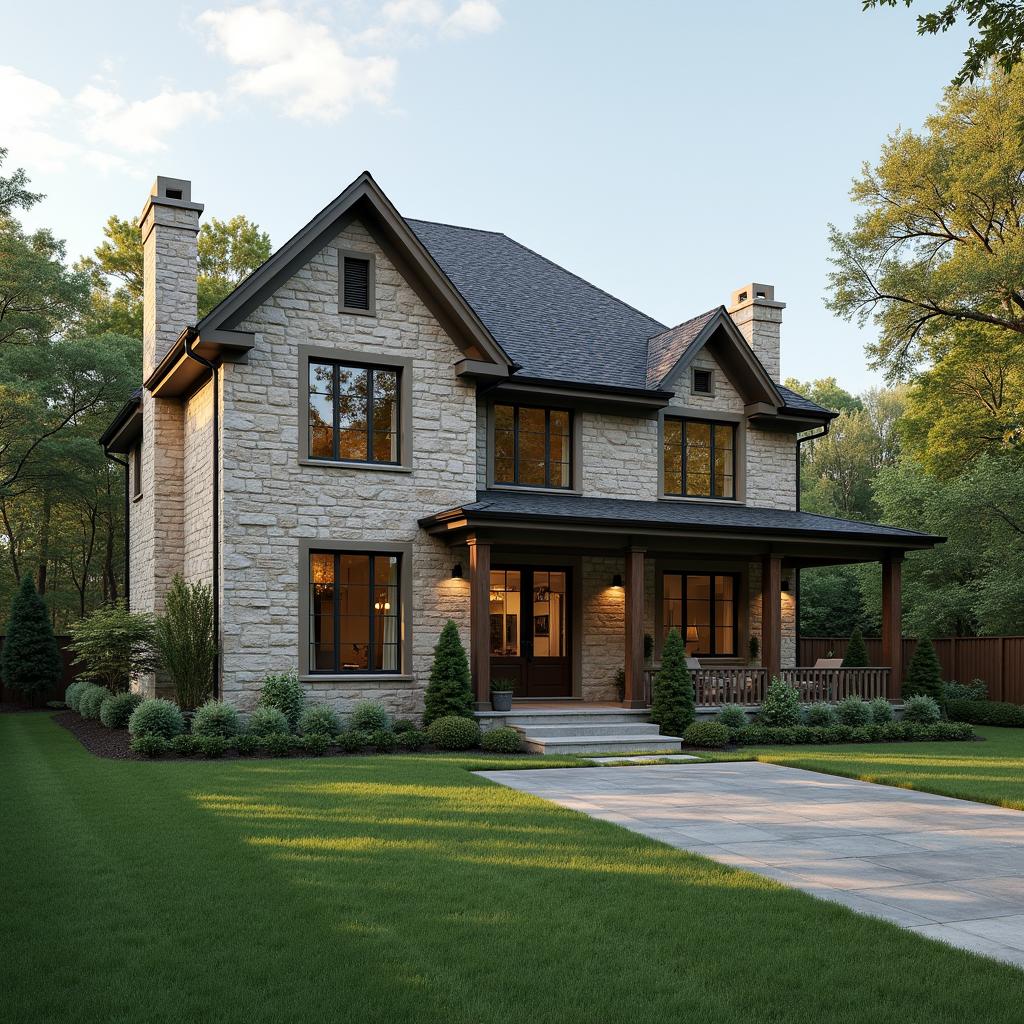} \\
        \rotatebox{90}{w/ \ours} & 
        \includegraphics[width=0.233\textwidth]{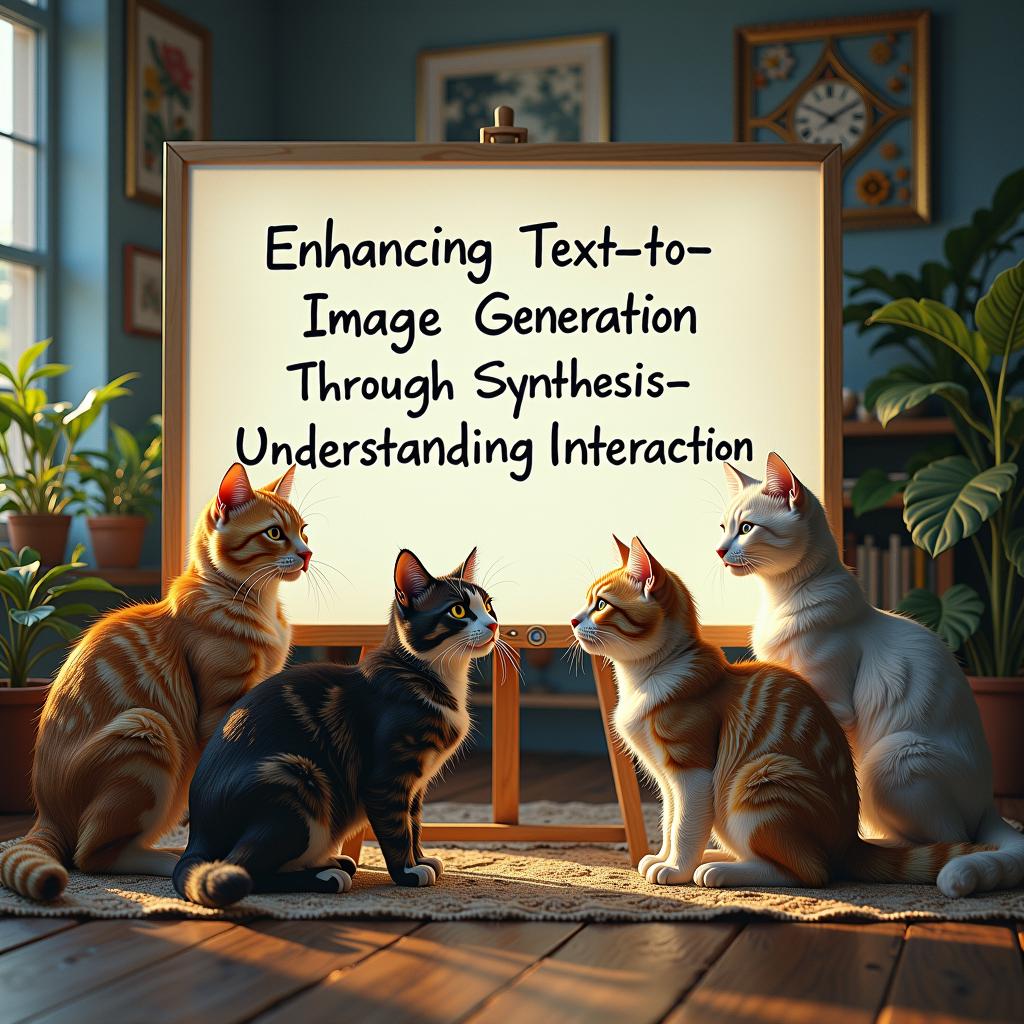} & 
        \includegraphics[width=0.233\textwidth]{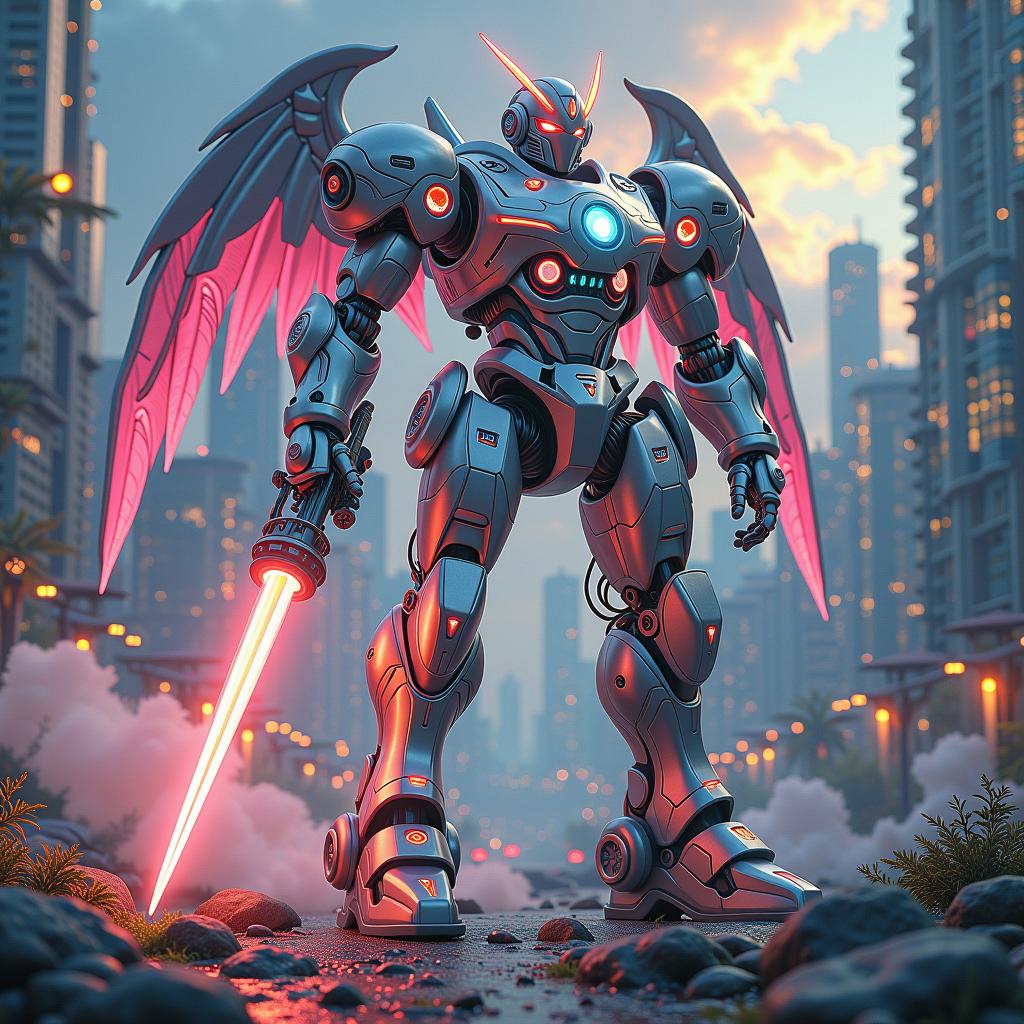} &
        \includegraphics[width=0.233\textwidth]{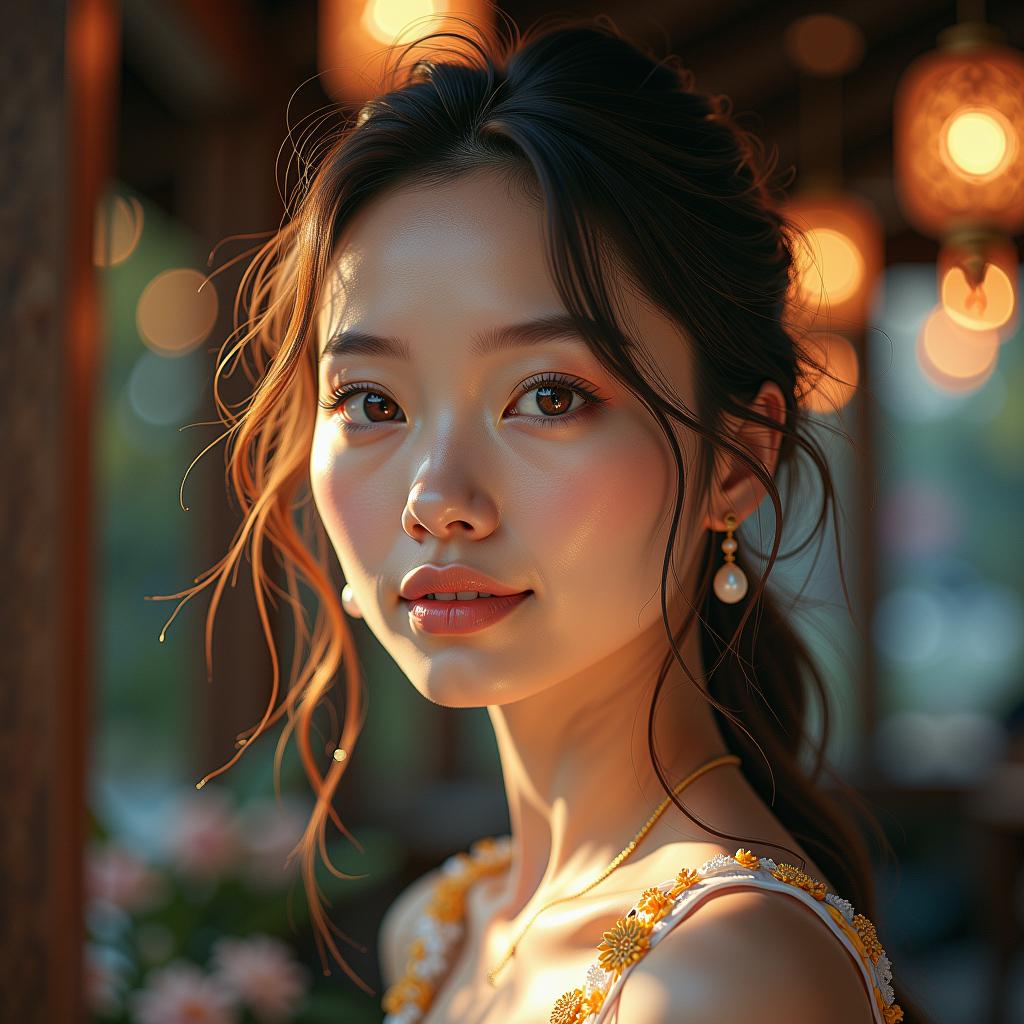} & 
        \includegraphics[width=0.233\textwidth]{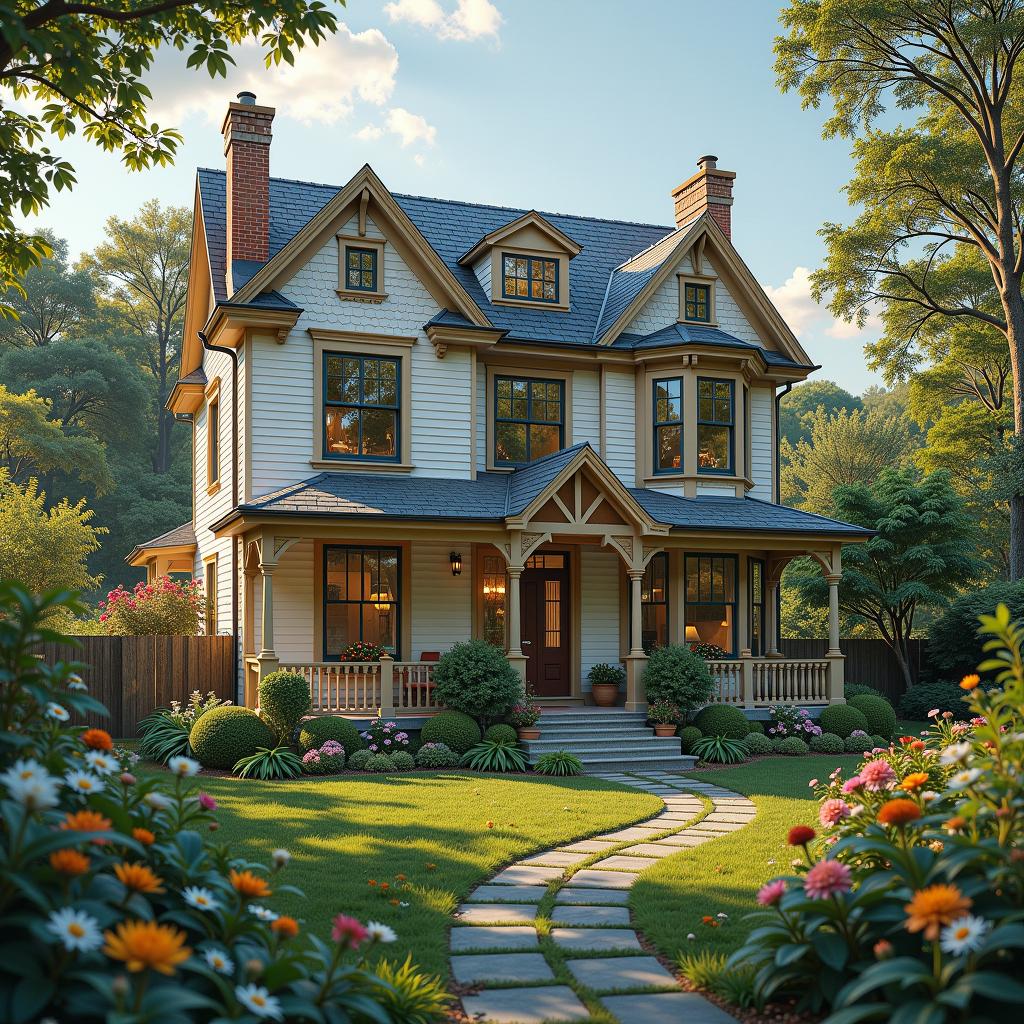} \\
    \end{tabular}
    \caption{Image examples improved by \ours. The base text-to-image model is FLUX.1[dev]. The \ours enhancement module is fused into the base model, without requiring additional computational resources.}
    \label{fig:gallery}
\end{figure*}

\begin{abstract}
The emergence of diffusion models has significantly advanced image synthesis. The recent studies of model interaction and self-corrective reasoning approach in large language models offer new insights for enhancing text-to-image models. Inspired by these studies, we propose a novel method called \ours for enhancing text-to-image models in this paper. To the best of our knowledge, \ours is the first one that improves image synthesis models via model interactions with understanding models. In the interactions, we leverage human preferences implicitly learned by image understanding models to provide fine-grained suggestions for image synthesis models. The interactions can modify the image content to make it aesthetically pleasing, such as adjusting exposure, changing shooting angles, and adding atmospheric effects. The enhancements brought by the interaction are iteratively fused into the synthesis model itself through an additional enhancement module. This enables the synthesis model to directly produce aesthetically pleasing images without any extra computational cost. In the experiments, we train the \ours enhancement module on existing text-to-image models. Various evaluation metrics consistently demonstrate that \ours enhances the generative capabilities of text-to-image models without incurring additional computational costs. The source code and models will be released publicly.

\end{abstract}

\section{Introduction}

\daoyuan{Considering the recent popular topic ``Reinforcement Fine-Tuning'' via small dataset. We may explicitly highlight our feature with a new title such as ``\ours: Interactively Fine-tuning Text-to-Image Models via Synthetic Data Rewarding''}

Diffusion models \cite{sohl2015deep, ho2020denoising} have been extensively studied in recent years. With the development of large-scale text-image datasets \cite{schuhmann2022laion, gu2022wukong}, pre-trained large text-to-image models \cite{rombach2022high, chen2023pixart, saharia2022photorealistic} have fully developed and demonstrated strong application potential. Downstream tasks such as interactive creation \cite{liu2024magicquill}, controllable image generation \cite{zhang2023adding}, and consistent story generation \cite{zhou2024storydiffusion} all require the generated content to align with human preferences. However, due to the training datasets being filled with a large number of low-quality images, pre-trained text-to-image models cannot generate high-quality images without guidance. This has become a challenge in the application of diffusion models.

To guide image generation models in producing high-quality images, current research primarily focuses on three aspects: \textbf{1) Data refinement} \cite{chen2024datajuicer, schuhmann2022laion} are employed to eliminate low-quality images from large training datasets, thereby preventing them from negatively impacting the model's performance. \textbf{2) Prompt engineering} \cite{wang2024discrete, cao2023beautifulprompt} aims to craft detailed prompts to guide the model in producing superior-quality images. \textbf{3) Alignment training} \cite{wallace2024diffusion, fan2024reinforcement} focuses on aligning the model's generative inclinations with human preferences via training. However, these methods all have certain limitations. Data refinement can only be used for coarse filtering. Directly filtering out low-quality images requires meticulous
efforts and potentially leads to overfitting due to the insufficient amount of data. Prompt engineering based on language models might result in generated images containing content that is inconsistent with the user-provided prompts, thereby compromising the text-image correlation. Alignment training is currently the key method for improving image quality. The mainstream alignment training methods, including Reinforcement Learning from Human Feedback (RLHF) \cite{ouyang2022training} and Direct Preference Optimization (DPO) \cite{rafailov2024direct}, require a large amount of manually annotated data, leading to extremely high costs.

On the other hand, recent studies of model interaction and self-corrective reasoning provide us with new insights for enhancing the capabilities of image generation models. Particularly, GPT-o1 \cite{gpto1} significantly enhances the capabilities of LLMs (Large Language Models) \cite{brown2020language} through self-corrective reasoning via the model itself, at the expense of longer computation time. LLMs are trained on human-generated data and potentially understand human interpretations and preferences for aesthetics. Recent studies have preliminarily demonstrated the feasibility of guiding image generation models through interactive conversations using language models \cite{huang2024dialoggen}. Some multimodal models \cite{wang2024qwen2, chen2024internvl, liu2024visual} are capable of understanding image content and expressing it through natural language, motivating us to explore the deeper assistive roles of LLMs in relation to image generation models.

To address the current challenges faced by image generation models, inspired by the model interaction and self-corrective reasoning approaches, we propose a novel text-to-image generation model enhancement approach called \ours. As shown in Figure \ref{fig:gallery}, \ours can significantly improve the image quality, aligning the generated image content with human preference.

The framework \ours is presented in Figure \ref{fig:workflow}. There are three modules in \ours, including a \textbf{generation module} for text-to-image generation, an \textbf{understanding module} for analyzing and refining the image content, and an \textbf{enhancement module} for improving the generation module. Firstly, we design an interaction algorithm, utilizing the understanding module to provide fine-grained modification suggestions for the generation module, and subsequently generate enhanced images. Secondly, we generate a large batch of image pairs using the interaction algorithm and filter them. Thirdly, we propose differential training, aimed at training the enhancement module to learn the differences before and after image enhancement. Fourthly, the enhancement module is then fused to the generation module, imbuing the generation module with the enhancement capability brought by interactions without requiring additional computational resources. By repeating the aforementioned process, we iteratively improve the generation module. In our experiments, we train the enhancement module for the state-of-the-art text-to-image model. \ours can notably enhance the performance of text-to-image models, resulting in the generation of aesthetically pleasing images. This improvement is consistently evidenced by various evaluation metrics. We will release the source codes\footnote{\url{https://github.com/modelscope/DiffSynth-Studio}} and the improved model\footnote{\url{https://www.modelscope.cn/models/DiffSynth-Studio/ArtAug-lora-FLUX.1dev-v1}}. Overall, the contributions of this paper include:
\begin{itemize}
    \item We design an interaction algorithm between a generation module and an understanding model in image synthesis, demonstrating that current multimodal LLMs can guide text-to-image models to generate high-quality images aligned with human preferences.
    \item We propose \ours, a framework for improving text-to-image models. By learning the differences between images before and after interaction, we iteratively enhance the capabilities of the text-to-image model.
    \item We train the \ours enhancement module based on the current state-of-the-art text-to-image models. Extensive experiments consistently demonstrate the effectiveness of \ours in improving image quality across multiple aspects.
\end{itemize}

        
        
        
        
        


\section{Related Work}

\subsection{Large Image Synthesis Models}

In recent years, diffusion models \cite{sohl2015deep, ho2020denoising} have achieved significant breakthroughs in the field of image synthesis. Since the introduction of Latent Diffusion \cite{rombach2022high}, models pre-trained on large-scale text-image datasets \cite{schuhmann2022laion, lin2014microsoft, gu2022wukong} have made considerable advancements. The generative capabilities of these models have been steadily improved, including both UNet-based models \cite{ronneberger2015u, rombach2022high, podell2023sdxl, sauer2025adversarial} and the more recent DiT-based models \cite{li2024hunyuan, chen2023pixart, esser2024scaling, flux}. Notably, DiT (Diffusion Transformer) \cite{peebles2023scalable} has considerably enhanced both the convergence speed and the generalization ability of image generation models, establishing itself as one of the most popular architectures in the realm of image synthesis. To further enhance image quality in terms of text-image alignment and aesthetic appeal, various approaches, such as data refinement \cite{chen2024datajuicer, schuhmann2022laion}, prompt engineering \cite{wang2024discrete, cao2023beautifulprompt}, and alignment training \cite{wallace2024diffusion, fan2024reinforcement}, have been extensively investigated. Based on these studies, we propose a new approach to enhance image quality in this paper.


\subsection{Aligning Models with Human Preferences}

\begin{figure}[]
    \centering
    \includegraphics[width=0.99\linewidth]{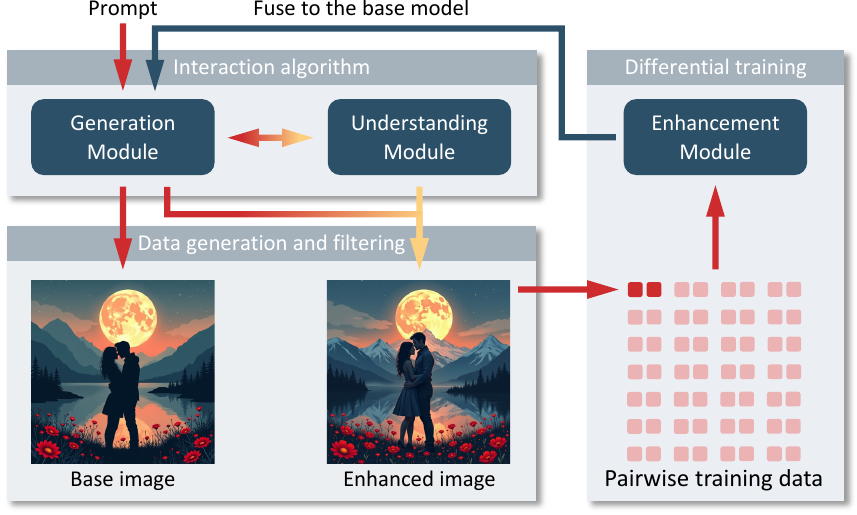}
    \caption{The framework of \ours encompasses three key components: the interaction algorithm, data generation and filtering, and differential training. This enhancement process can be iteratively applied to the model, facilitating iterative improvement. \daoyuan{Consider making the figure more technical? e.g., the module is too abstract, or we may refer to a pseudocode in Appendix (with more steps and notations).}}
    \label{fig:workflow}
\end{figure}

\begin{figure*}
  \includegraphics[width=0.99\linewidth]{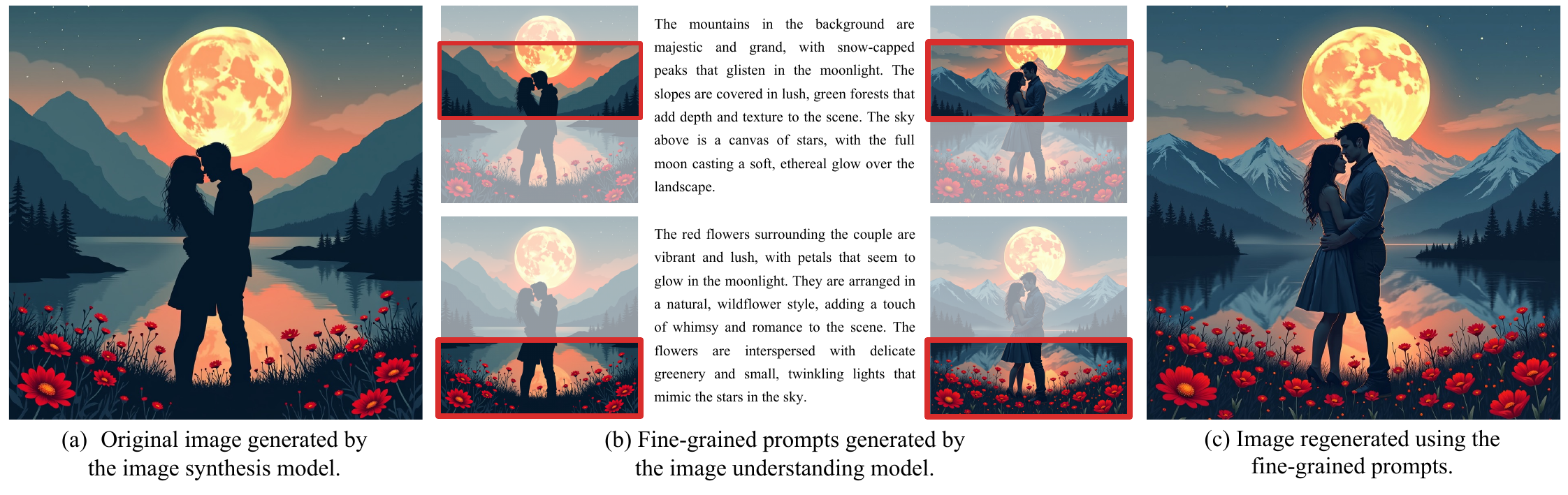}
  \caption{An example of the synthesis-understanding interaction process. By utilizing image understanding models to analyze and generate fine-grained prompts, we can enhance the overall quality of generated images.}
  \label{figure:interaction}
\end{figure*}

Text-to-image models pre-trained on extensive text-image datasets have demonstrated rudimentary image generation abilities, but these models often produce suboptimal quality images without fine-tuning \cite{liu2024alignment}. Currently, alignment training stands as the principal method for improving image quality by aligning generated content with human preferences. Alignment training is initially investigated in large language models \cite{ouyang2022training, rafailov2024direct}, and has recently been applied to diffusion models. For example, based on reinforcement learning, approaches like DPOK \cite{fan2024reinforcement} and DDPO \cite{black2023training} gather human preferences on model-generated outputs for fine-tuning text-to-image models. Similarly, Diffusion-DPO \cite{wallace2024diffusion} and SPO \cite{liang2024step} employ auxiliary models to model human preferences, using DPO \cite{rafailov2024direct} to fine-tune diffusion models accordingly. However, because human preferences are difficult to quantify, these alignment training methodologies necessitate extensive manually labeled datasets, which are prohibitively expensive to produce. Inspired by these studies, we explore the possibility of using multimodal LLMs to replace manual annotation, aiming to obtain a large amount of training data at a lower cost for alignment training.

\section{Methodology}

The framework of \ours is presented in Figure \ref{fig:workflow}. \ours consists of three key components: the interaction algorithm, data generation and filtering, and differential training. In this section, we provide a detailed description of each module.


\subsection{Interaction Algorithm}
\label{sec:interaction}

Empirically, we observe that text-to-image models tend to generate simple content when given simple prompts. Prompt engineering is generally essential for generating high-quality images, but crafting high-quality prompts manually poses certain challenges. Therefore, a single image synthesis module (a text-to-image model) struggles to generate detailed and aesthetically pleasing images. To address this challenge, we propose an interactive algorithm and utilize an additional understanding module to aid the synthesis module.

The interactive algorithm includes three steps: generation, understanding, and refinement. First, we use the original generation pipeline of the text-to-image model to generate an image $X$. Second, we employ the understanding module $u$ to analyze the image content and generate modification suggestions. The understanding module is implemented based on multimodal LLMs due to their significant image understanding and grounding capabilities. The modification suggestions provided by the understanding module are in the form of $n$ pairs of bounding box and prompt, which are formulated as $u(X)=\{(P_i,M_i)\}_{i=1}^n$, where the bounding box $M_i\in\{0,1\}^{H\times W}$ represents the location and the prompt $P_i$ describes the corresponding modified content. To improve computational efficiency, we directly generate all bounding boxes and prompts through a single-turn dialogue. Third, we use the synthesis module to regenerate the image according to the suggestions for image modifications. To finely control the content of images and ensure that each prompt affects the corresponding area, we design a partitioned image generation method based on previous studies \cite{li2023gligen, bar2023multidiffusion}. Assuming that the original model output is $\hat\epsilon_{\theta} (P,t,h)$, where $P$ is the original prompt, $t$ is the timestep of the denoising process, and $h\in \mathbb R^{H\times W}$ is the latent representation of the image, we use
\begin{align*}
    &\hat \epsilon_{\theta} (P,t,h|\{(P_i,M_i)\}_{i=1}^n)\\
    =&\frac{\hat\epsilon_{\theta} (P,t,h) + \sum_{i=1}^n \hat\epsilon_{\theta}(P_i,t,h)\cdot M_i}{1+\sum_{i=1}^n M_i}
\end{align*}
to replace the original model output, where $\cdot$ denotes the element-wise multiplication. Intuitively, the interaction algorithm employs the model to infer according to different prompts and then performs a weighted average of the results based on the location information.


\subsection{Data Generation and Filtering}
\label{sec:filter}

Although the procedures described in the last subsection can improve the quality of generated images without training, this algorithm has two drawbacks. 1) Slow computational speed. Since fine-grained prompts require independent forward inferences through the model, the overall computation time increases to $n+1$ times the original. This prolongs the generation process to several minutes. 2) Risk of bad cases. Due to the lack of end-to-end training in the modified image generation pipeline, there is a potential for producing unrealistic images. To overcome these drawbacks, we aim to consolidate the enhancements produced through interaction into the text-to-image model itself via post-training. Specifically, we need to generate a large batch of image pairs $(X,X')$ through interaction, where $X$ represents images generated by the original text-to-image pipeline and $X'$ represents the interactively refined images. After filtering, a smaller dataset is obtained, which is then used for training.

In the data filtering process, we apply stringent filtering criteria. Fundamentally, we expect that images enhanced through interaction should be more aesthetically pleasing than the originals. Therefore, we only retain image pairs with increased aesthetic scores \cite{schuhmann2022laion}. Additionally, we need to ensure that the enhanced images are consistent with the semantic meaning of the text prompts. To achieve this, we use the CLIP model \cite{radford2021learning} to filter out all image pairs where the text-image similarity decreases. These two filtering steps can effectively eliminate a significant portion of the data unsuitable for training. However, given that the prompts are collected from real users and exhibit complex and varied content, we conduct further meticulous manual reviews. During the manual review, we remove the content related to pornography, violence, politics, and racial discrimination, and ensure that the enhanced images in the image pairs show a clear improvement in quality. Appendix \ref{app:examples} shows several categories of the final retained training data, where the image pairs exhibit significant improvements in various aspects.

\subsection{Differential Training}
\label{sec:train}

After filtering and review, approximately 1\% to 2\% images are retained. The amount of data is relatively small, and directly using these data to fine-tune the model leads to overfitting. As shown in our preliminary experiments. Based on the observation, we propose a differential training approach to learn the differences between images, rather than directly learning the images enhanced by interactions.


The model structure of the enhancement module is LoRA \cite{hu2021lora}. Assume that the parameters of fully connected layers in the original model are formulated as
\begin{equation*}
    \theta = \{W_i\}_{i=1}^m.
\end{equation*}
For each parameter matrix $W_i\in\mathbb R^{d_1\times d_2}$, the parameter matrix after adding LoRA becomes $W_i+ B_iA_i$, where $A_i\in \mathbb R^{r\times d_2}$ and $B_i\in \mathbb R^{d_1\times r}$ are two low-rank matrices. The LoRA rank $r$ is a hyperparameter. We use $\phi=\{(A_i,B_i)\}_{i=1}^m$ to denote all LoRA parameters and use ``$\oplus$'' to denote the adding LoRA operator, i.e.,
\begin{equation*}
    \theta \oplus \phi = \{W_i+B_iA_i\}_{i=1}^m.
\end{equation*}
For a diffusion model based on flow matching \cite{esser2024scaling, flux}, the LoRA training problem using a single image $X$ with its latent representation $x$ is formulated as
\begin{equation*}
    \Phi(\theta, X) = \mathop{\arg\min}_{\phi} \mathop{\mathbb E}_{t\sim \mathcal T, \epsilon\sim \mathcal N(0,1)}\left[\mathcal L(P,t,h,\epsilon)\right],
\end{equation*}
where
\begin{equation*}
    \mathcal L(P,t,h,\epsilon)=w_t||\hat \epsilon_{\theta\oplus\phi}(P,t,h_t)-\epsilon+h||_2^2,
\end{equation*}
\begin{equation*}
    h_t=(1-\sigma_t)h+\sigma_t \epsilon.
\end{equation*}
For a denoising diffusion probabilistic model \cite{rombach2022high, podell2023sdxl}, the LoRA training problem can be similarly formulated. We only need to replace the loss function as
\begin{equation*}
    \mathcal L(P,t,h,\epsilon)=||\hat \epsilon_{\theta\oplus\phi}(P,t,h_t)-\epsilon||_2^2,
\end{equation*}
where
\begin{equation*}
    h_t=\sqrt{1-\alpha_t}h+\sqrt{\alpha_t}\epsilon.
\end{equation*}
The parameters $w_t$, $\sigma_t$, $\alpha_t$, and the probabilistic distribution $\mathcal T$ are determined by the flow matching theory and the probabilistic diffusion process. We recommend readers to read the corresponding original papers \cite{liu2022flow, ho2020denoising} for more details. Our training approach is compatible with both two kinds of diffusion models.

\begin{table*}[htbp]
\caption{Quantitative results of \ours on existing text-to-image models.}
\label{tab:quantitative}
\centering
\setlength{\tabcolsep}{9pt}{
\begin{tabular}{l|cccc|c|c}
\hline
                         & \multicolumn{4}{c|}{Image quality}                                             & Human evaluation    & Ethtic metric        \\ \cline{2-7}
                         & \makebox[0.07\textwidth][c]{Aesthetic $\uparrow$} & \makebox[0.07\textwidth][c]{PickScore $\uparrow$} & \makebox[0.07\textwidth][c]{MPS $\uparrow$} & \makebox[0.07\textwidth][c]{CLIP $\uparrow$} & \makebox[0.07\textwidth][c]{Win rate $\uparrow$} & \makebox[0.07\textwidth][c]{NudeNet $\downarrow$} \\ \hline
FLUX.1{[}dev{]}          & 6.35                 & 42.22                & 47.52          & 26.92           & 39.18               & 6.19                 \\
FLUX.1{[}dev{]} + \ours & \textbf{6.81}        & \textbf{57.78}       & \textbf{52.48} & \textbf{26.97}  & \textbf{45.93}      & \textbf{4.44}        \\ \hline
\end{tabular}}
\end{table*}

Given an image pair $(X,X')$ generated by the interaction algorithm. We train the first LoRA model $\phi_1=\Phi(\theta,X)$ using the single image generated by the original model so that it always generates image $X$ when inputting the corresponding prompt. Then, based on this, we train the second LoRA model $\phi_2=\Phi(\theta\oplus\phi_1,X')=\Phi\big(\theta\oplus\Phi(\theta,X),X'\big)$ so that it always generates image $X'$ when inputting the corresponding prompt. Thus, the second LoRA $\phi_2$ can be used to represent the differences between the two images. We drop the first LoRA and call the second LoRA \textbf{differential LoRA}. We train a corresponding differential LoRA model for each image pair. The training process for each image pair is independent, which makes our training process distributed and scalable.

\subsection{Iterative Improvement}

Through differential training, we obtain a LoRA model $\phi$ that can enhance the generative capabilities of the text-to-image model. This LoRA model can be fused into the base model, i.e., let
\begin{equation*}
    \theta \leftarrow \theta \oplus \alpha \Phi\big(\theta\oplus\Phi(\theta,X),X'\big),
\end{equation*}
where $\alpha$ is the weight of the LoRA. To enhance the stability of the model's preferences, we average the LoRA parameters across multiple image pairs. Based on this iterative formula, we make it possible to continue generating data through the interaction processes. Consequently, the data generation and the differential training process can be iteratively repeated until the interactive algorithm can no longer significantly improve the quality of the generated images. Ultimately, we obtain a series of stacked LoRA models $\{\phi^{[1]},\phi^{[2]},\dots\}$. We merge them into a single LoRA model by concatenating the corresponding matrices. The use of the entire enhancement module is consistent with that of a standard LoRA model and maintains compatibility with other LoRA models. Additionally, users can adjust the influence of the enhancement module on the text-to-image model by tuning the weight of the merged LoRA model, thereby achieving controllable generation. 

From another perspective, this iterative enhancement process involves updating the model parameters at each iteration, akin to a gradient descent step. We provide a detailed analysis of the iterations in the experiments in Section \ref{sec:iteration_experiment}. The trainable LoRA parameters correspond to the gradient. The parameter $\alpha$ corresponds to the learning rate in gradient descent. A smaller $\alpha$ can make the training process more stable, but it will slow down the convergence speed. The number of averaged LoRA models corresponds to the batch size. In this manner, we can employ human preference, an inherently non-differentiable training objective, for the training of the model implicitly via data synthesis.

\daoyuan{Overall Suggestions on the Methodology part:
\begin{enumerate}
    \item Discuss the role and impact of iteration steps and the $\alpha$. Besides, you can simply cross-refer to subsequent experiments.
    \item Elaborate on the choice and limitations of the image generation module and understanding module. Discuss how these choices impact the overall algorithm's performance in a highlighting manner (current version mixed many details, easily confusing readers). For example: (a) What are the criteria for model selection? (b) What are the strengths and weaknesses of the chosen models? (c) How does enhancing one module over the other affect the results? (d) What are the potential and expected outcomes of enhancing both modules?
    \item (Theoretical Analysis) Formalize the algorithm, potentially using a bi-level optimization framework. Then we may discuss concepts like Pareto optimality and convergence analysis, or explore what evidence exists regarding its optimality or convergence. This should be make the paper more ICML-style rather than CVPR-style.
\end{enumerate}
}

\section{Experiments}

We conduct extensive experiments to demonstrate \ours's effectiveness, including improving off-the-shelf models and thoroughly investigating each component.

\subsection{Improving Off-the-Shelf Models}

\subsubsection{Experimental Settings}

We train the \ours LoRA based on the state-of-the-art text-to-image model ``FLUX.1[dev]'' \cite{flux}. In the interaction algorithm, the understanding module is implemented based on Qwen2-VL-72B \cite{wang2024qwen2} due to its sufficiently accurate visual grounding capabilities, which enable the generation of fine-grained bounding boxes and prompts. The prompt used in Qwen2-VL-72B is presented in Appendix \ref{app:prompt}. We provide detailed discussions about the selection of the multimodal LLM in Appendix \ref{app:mllm}, including a comparative analysis between six SOTA multimodal LLMs. Our experiments do not require a text-image dataset; we only use a dataset of prompts. In each training iteration, we randomly sample approximately 3k prompts from the DiffusionDB dataset \cite{wang2022diffusiondb}. Considering that these prompts are collected from users on the internet and some may contain ambiguous semantics, we refine the prompts using Qwen2-VL-72B before generating images. After filtering and reviewing as described in Section \ref{sec:filter}, approximately 30 to 100 image pairs remain. We trained a differential LoRA model for each image pair. The learning rate is set to $1\times 10^{-4}$, with a batch size of 1, and the LoRA model is trained for 400 steps. The LoRA rank is manually adjusted to 4, 8, or 16 to ensure convergence on the training image. The loss function is consistent with the flow match theory, and other training hyperparameters are consistent with those of the FLUX.1[dev] model itself.

\subsubsection{Quantitative Comparison}

After training, we randomly sample 10k prompts in DiffusionDB \cite{wang2022diffusiondb} to evaluate the quality of the generated images. These prompts are not used in the data generation. The evaluation metrics include:
\begin{itemize}
    \item \textbf{Aesthetic} \cite{schuhmann2022laion}: This metric utilizes an aesthetic evaluation model trained on human-labeled LAION image data. The model assesses the visual appeal of the images and is widely used in data processing.
    \item \textbf{PickScore} \cite{kirstain2023pick}: A model trained on the Pick-a-Pic dataset based on human preferences for model-generated images. This model can compare two images and estimate the probability that humans consider one image to be of higher quality.
    \item \textbf{MPS} \cite{zhang2024learning}: A multi-dimensional human preference evaluation model. This model can comprehensively assess image-text consistency, aesthetics, and details, providing a score that represents the overall quality of the image.
    \item \textbf{CLIP} \cite{radford2021learning}: A multimodal model trained using contrastive learning, utilized in our experiments to calculate the similarity between text and image, which is a measure of text-image consistency.
\end{itemize}

The quantitative results are presented in Table \ref{tab:quantitative}. In the three aesthetic quality metrics, the model trained with \ours demonstrates a significant improvement. This clearly indicates the efficacy of \ours in enhancing image quality. Furthermore, the CLIP text-image similarity metric does not exhibit a noticeable decline, suggesting that \ours does not compromise the original text comprehension ability of the base model. Overall, \ours serves to enhance the fundamental capabilities of the text-to-image model, without sacrificing their linguistic interpretative abilities.

\subsubsection{Human Evaluation}

\begin{figure*}[htbp]
    \centering
    \includegraphics[width=0.99\textwidth]{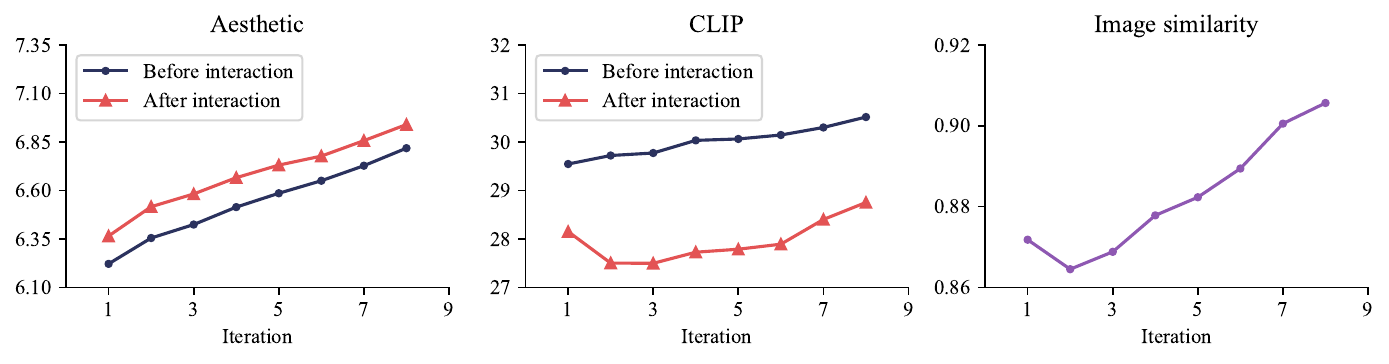}
    \caption{Statistical information of image pairs generated during the iterative training process. These statistical metrics are calculated based on the refined prompt utilized during the training process and are distinct from those presented in Table \ref{tab:quantitative}.}
    \label{fig:visualize_iteration}
\end{figure*}

Some studies \cite{podell2023sdxl, jiang2024genai} have highlighted the limitations of automatically computed evaluation metrics, prompting us to conduct an additional double-blind human evaluation. We invite 20 participants to take part in this evaluation. In each round, participants are shown two images: one generated by the original text-to-image model and the other generated by the \ours-enhanced model. Similar to GenAI-Arena \cite{jiang2024genai}, the positions of the two images are randomized. Each participant is asked to select the image with better visual quality or to choose ``tie''. We record the percentage of user votes, as shown in Table \ref{tab:quantitative}. \ours achieves a winning rate of 45.93\%, demonstrating the effectiveness of \ours in enhancing visual quality.

\subsubsection{Ethics Consideration}

Although \ours can enhance the capabilities of image generation models, this also introduces potential ethical risks. We have observed that user-provided prompts collected from the internet contain descriptions of harmful content such as pornography and violence. Pre-trained text-to-image models have the ability to generate such harmful content. Additionally, pre-trained image understanding models potentially learn biases towards pornographic content, resulting in suggestive content appearing in the images generated by the interaction algorithm. This necessitates additional human resources to review the generated datasets during the iterative training process of \ours, otherwise, these biases would be learned and propagated. We have tried to prevent the model from being influenced by these biases. To ensure that \ours does not increase the propensity of the model to generate harmful content, we conduct additional experiments. We use NudeNet \cite{bedapudi2019nudenet} to calculate the harmful scores of the images generated in the aforementioned experiments. The results, presented in the last column of Table \ref{tab:quantitative}, show that \ours does not increase the propensity of the model to generate harmful content.


\subsection{Impact of Iteration Step}
\label{sec:iteration_experiment}

To better understand the changes in the model's capabilities throughout the iterative training process, we analyzed the data of image pairs generated in each iteration. Some statistical indicators are presented in Figure \ref{fig:visualize_iteration}. In each iteration, our primary focus is on the enhancement of image aesthetics by the interaction algorithm. It can be observed from the figure that the aesthetic scores consistently improve after interaction. This enhancement capability is ingrained into the model during training and carries over to the next iteration, enabling continuous improvement. We also calculated the correlation between images and prompts before and after interaction using the CLIP model. It should be noted that these prompts are refined by the language model, so the CLIP scores appear slightly higher than those in Table \ref{tab:quantitative}. Although the interaction algorithm may sometimes alter the image content away from its original semantics, especially when the prompt contains terms like ugly, dirty, or bloody, our rigorous data filtering process eliminates such data to prevent compromising the model's original capabilities. Additionally, we calculated the cosine similarity of images before and after interaction using the vision encoder component \cite{dosovitskiy2020image} of the CLIP model. As iterations progress, the enhancement effect of the interaction algorithm on image quality diminishes. In the eighth iteration, we are unable to obtain sufficient image pairs for training after filtering, and thus, we stop the training process.

\daoyuan{So, if we have larger original dataset, the improvement can be scaled as well?}

\subsection{Ablation Studies}

\subsubsection{Interaction Algorithm}

\begin{figure}[]
    \centering
    \setlength{\tabcolsep}{3pt}
    \begin{tabular}{ccc}

        \includegraphics[width=0.31\linewidth]{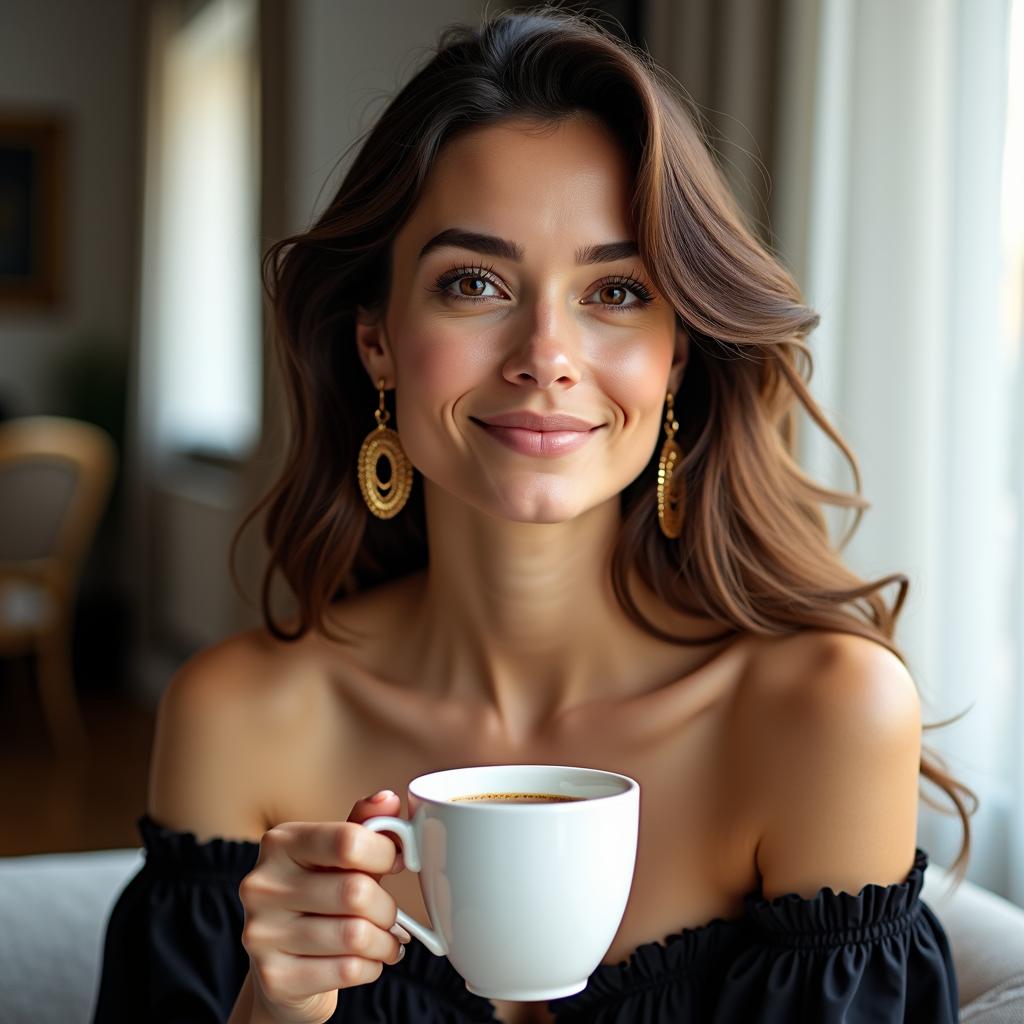} & 
        \includegraphics[width=0.31\linewidth]{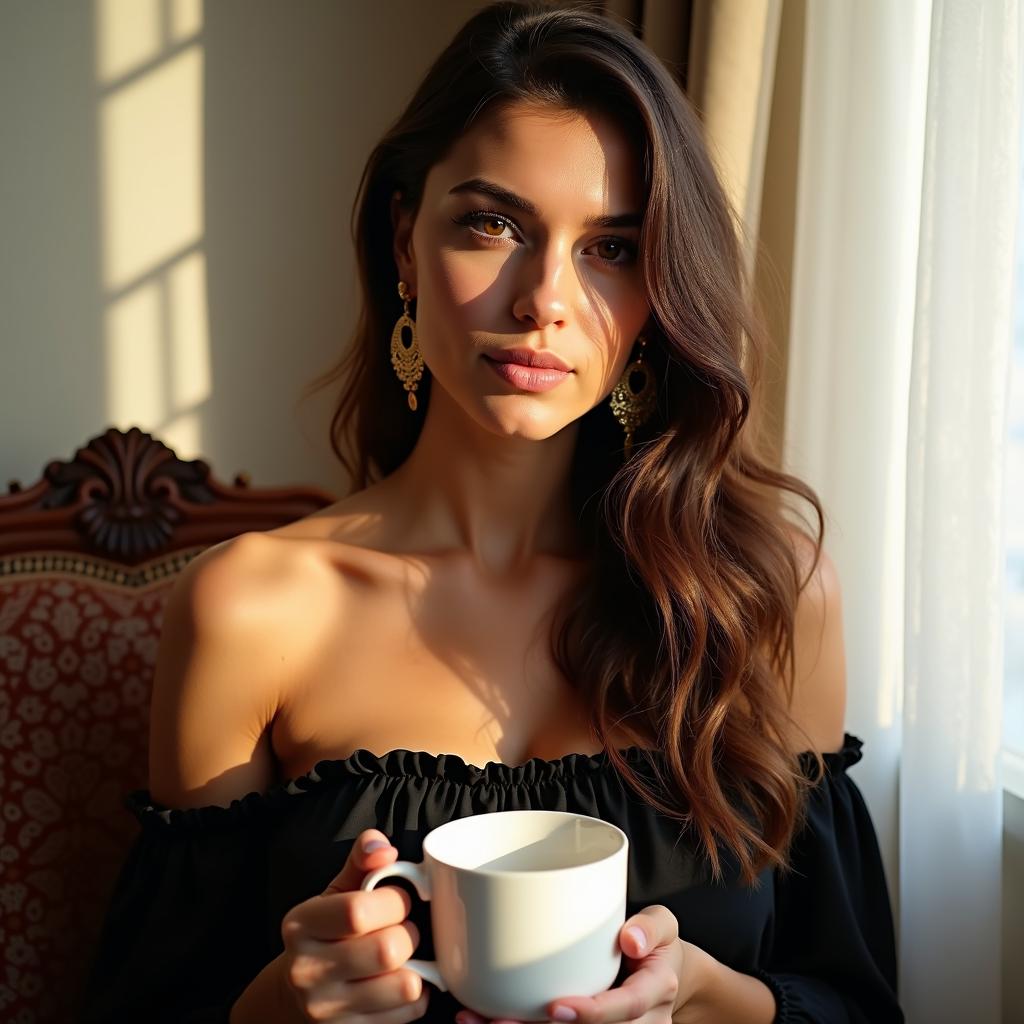} &
        \includegraphics[width=0.31\linewidth]{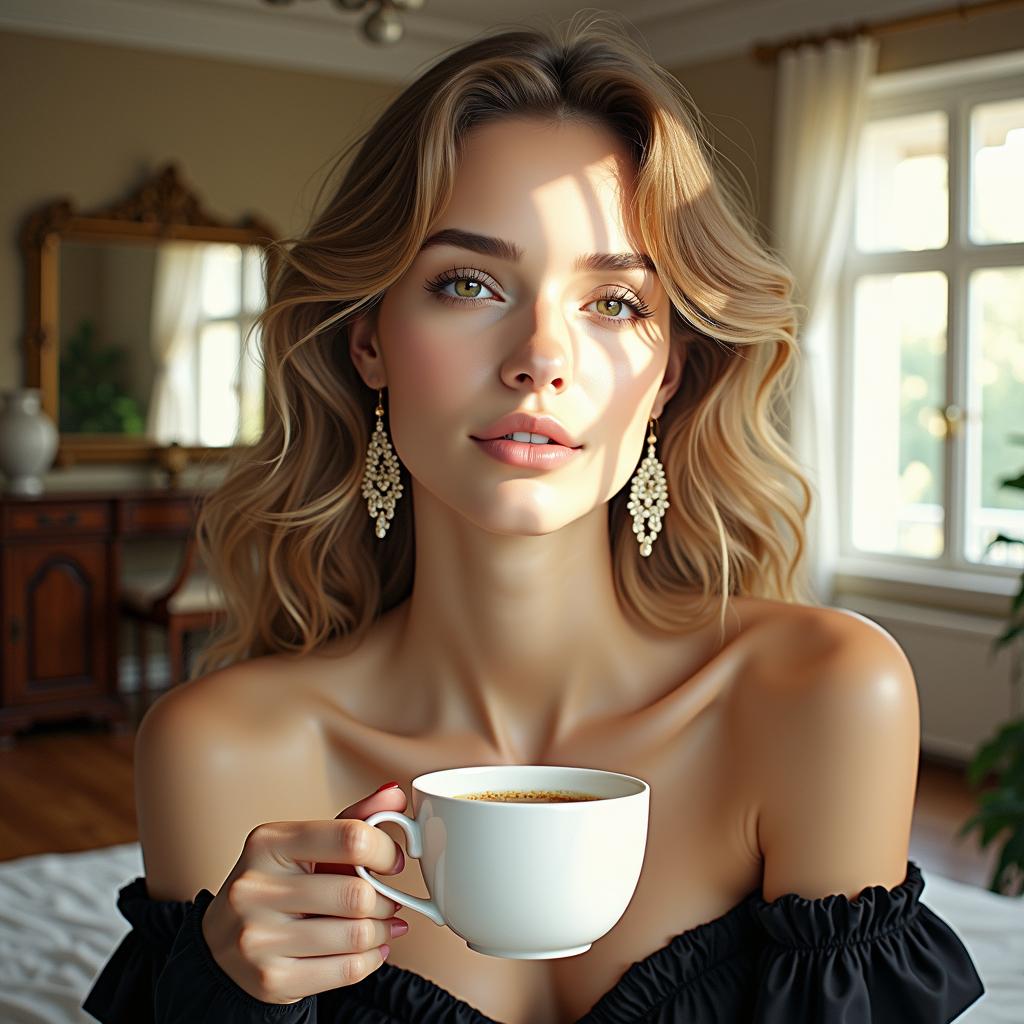} \\

        \includegraphics[width=0.31\linewidth]{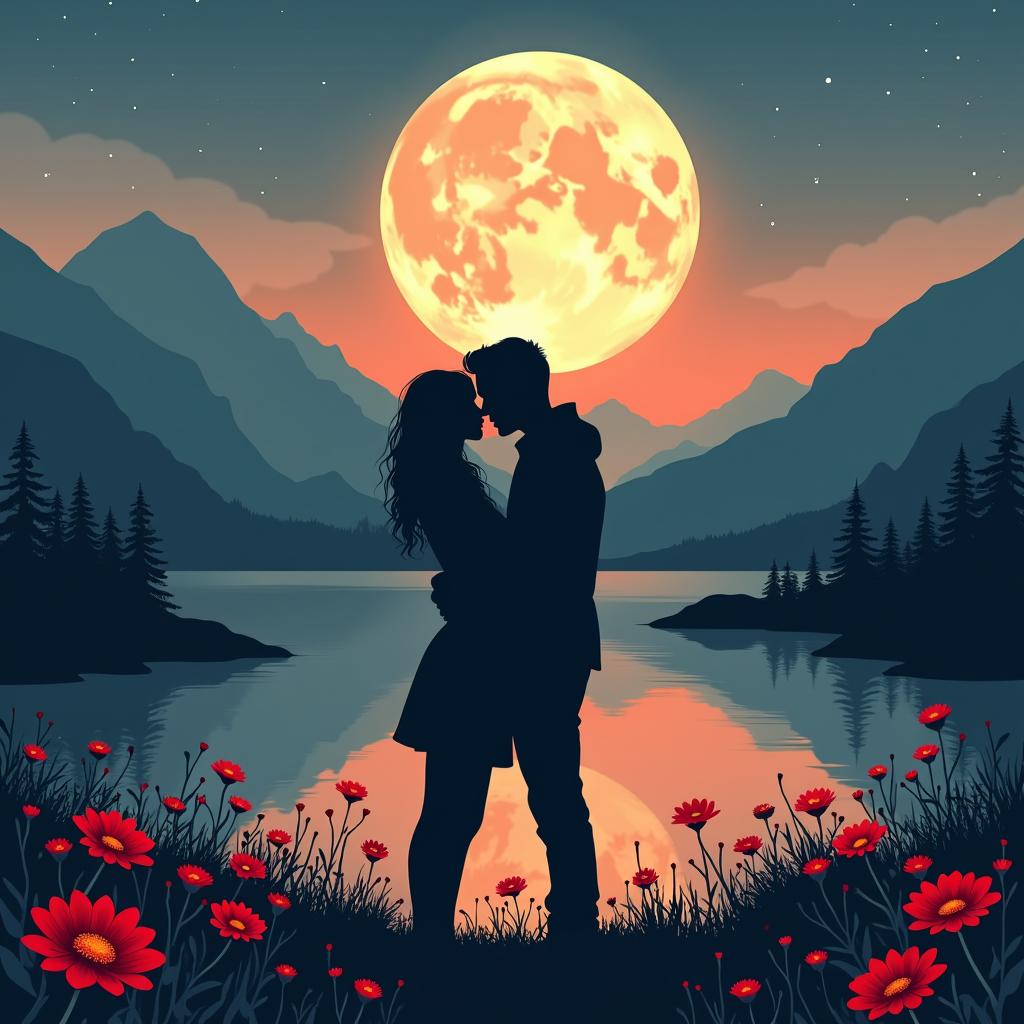} & 
        \includegraphics[width=0.31\linewidth]{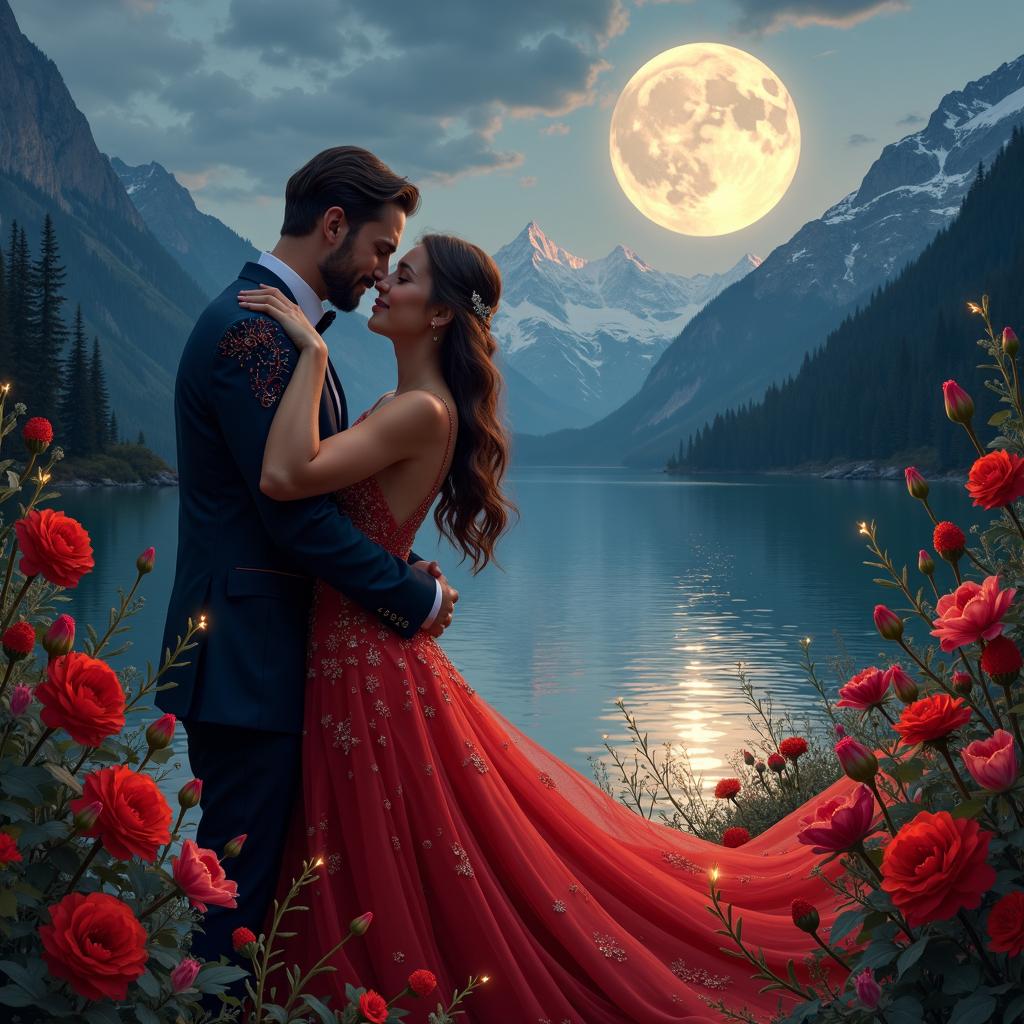} &
        \includegraphics[width=0.31\linewidth]{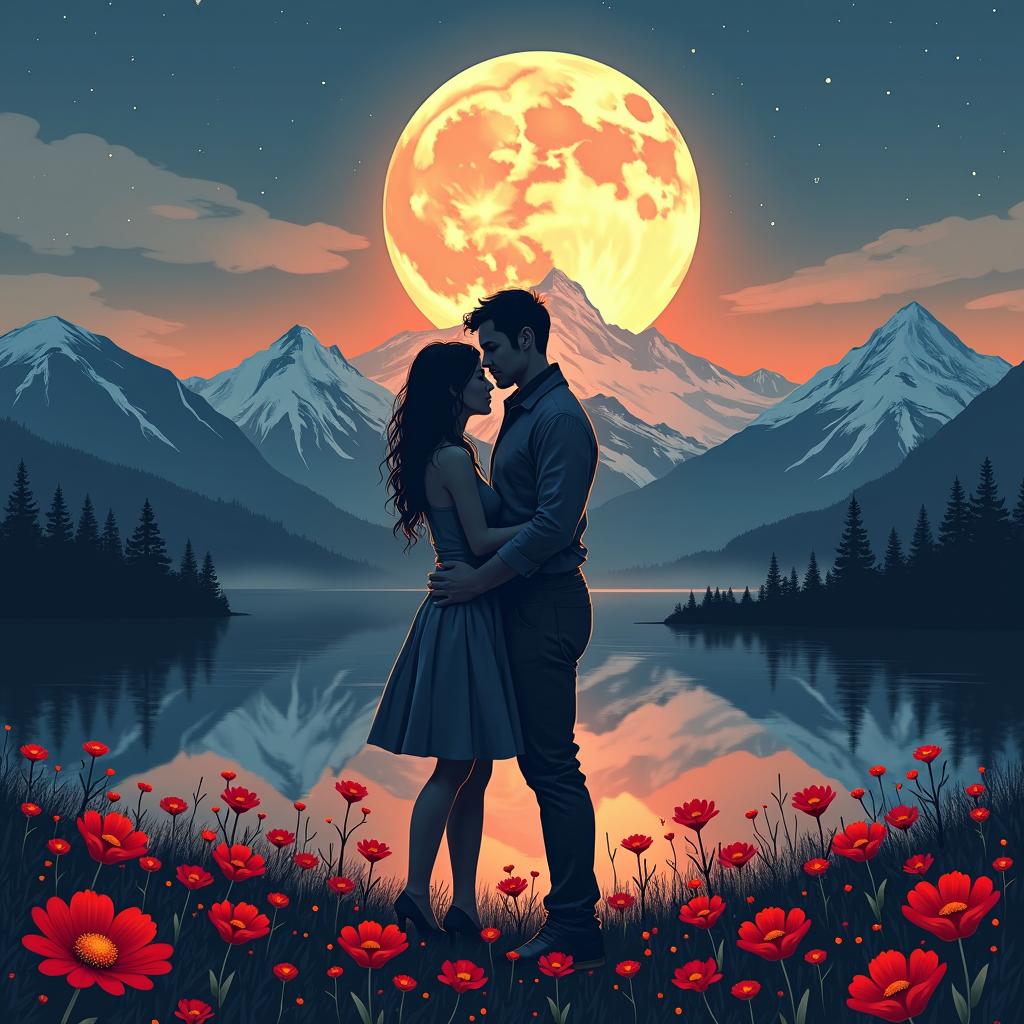} \\
        
        \includegraphics[width=0.31\linewidth]{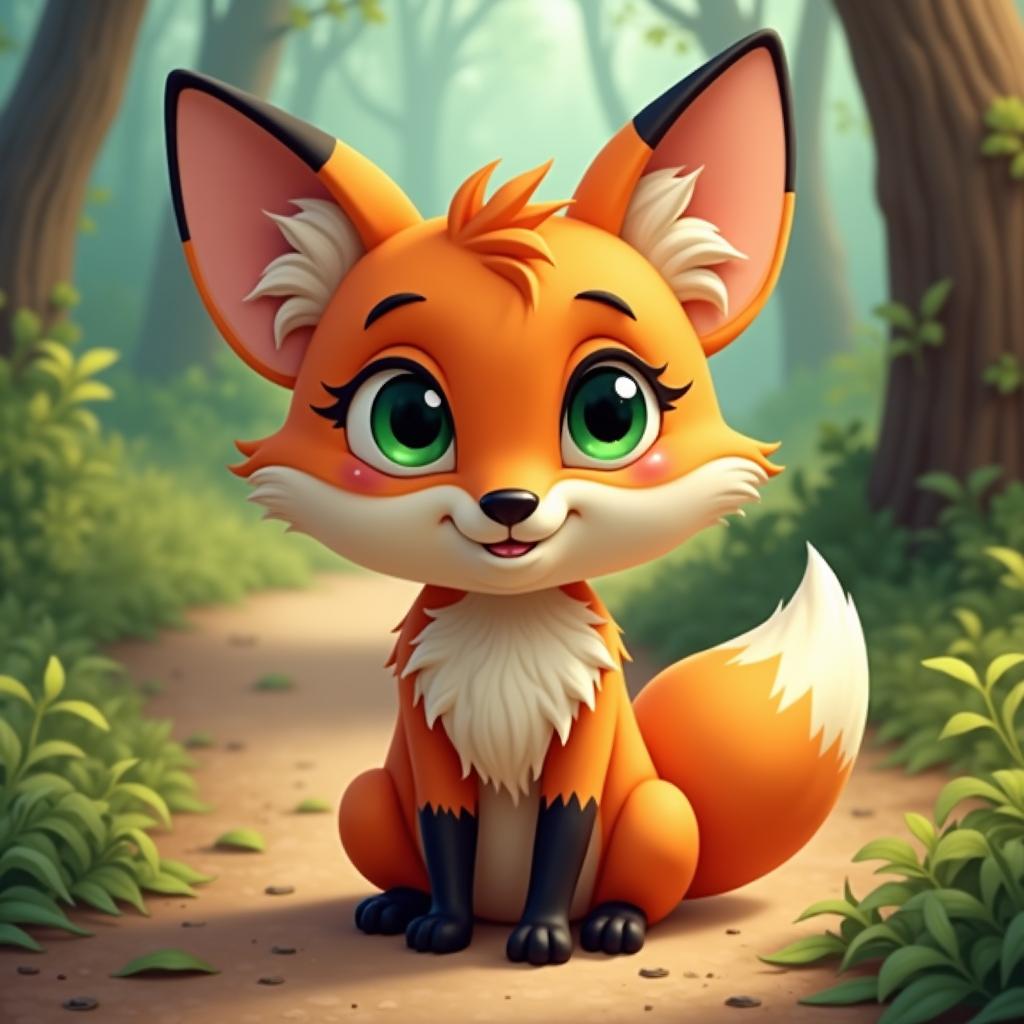} & 
        \includegraphics[width=0.31\linewidth]{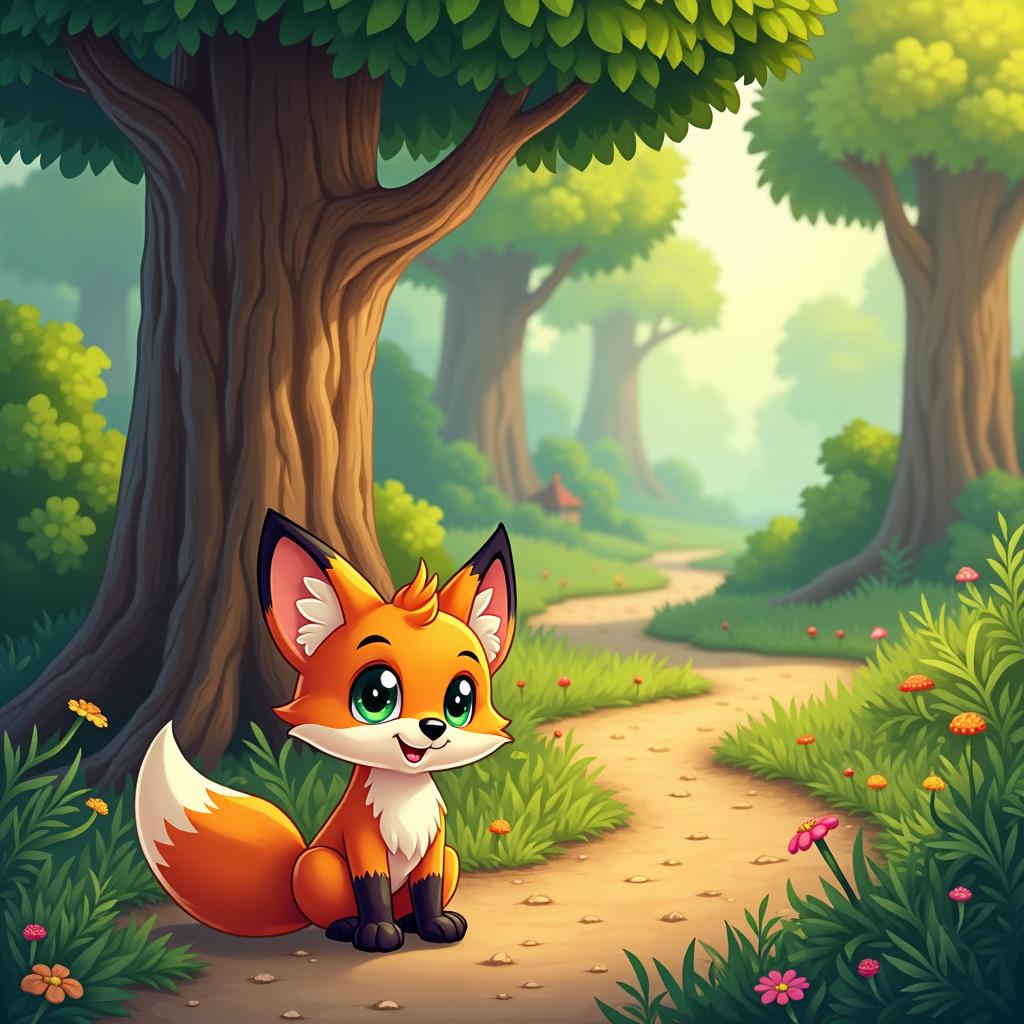} &
        \includegraphics[width=0.31\linewidth]{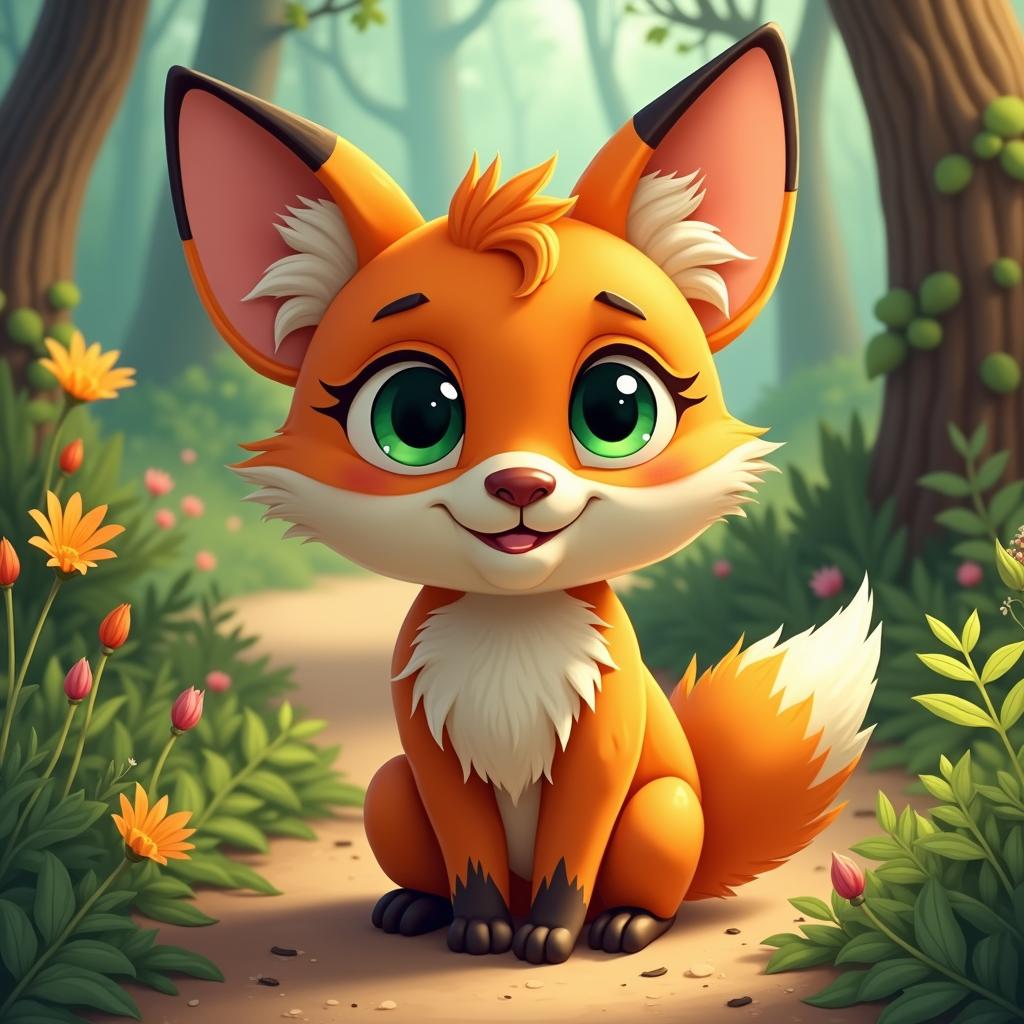} \\
        
        \scriptsize{(a) Image generated by} & \scriptsize{(b) Image generated by} & \scriptsize{(c) Image generated by} \\
        \scriptsize{the original prompt.} & \scriptsize{the refined prompt.} & \scriptsize{our interaction algorithm.} \\

    \end{tabular}
    \caption{Comparison of naive prompt refining and our interaction algorithm.}
    \label{fig:ablation_interaction}
\end{figure}

We compared the interaction algorithm outlined in Section \ref{sec:interaction} with the naive prompt refinement approach, as demonstrated in Figure \ref{fig:ablation_interaction}. In the naive prompt refinement approach, we leverage the multimodal LLM to directly generate a detailed prompt for the image generation model. This example reveals that regenerating images using only refined prompts typically results in a complete alteration of the scene's overall composition. Conversely, our interaction algorithm is capable of enhancing the details while preserving the fundamental composition, exemplified by the flowers and light in the image. This suggests that our interactive algorithm effectively utilizes the capabilities of the image understanding model to generate fine-grained prompts ensuring consistency in image content. Therefore, the image pairs generated using the interaction algorithm are better suited for the subsequent differential training process, which is aimed at learning the differences between two images.

\subsubsection{Differential Training}

We also investigate the effectiveness of the differential LoRA training mentioned in Section \ref{sec:train}. We compare it with a LoRA model naively trained using the filtered data. We evaluate the two training methods in the first iteration of FLUX.1[dev]. By employing the same learning rate and number of training steps, we calculate the four evaluation metrics of the LoRA models. As shown in Table \ref{tab:differential_training}, it is evident that naive LoRA training leads to significant overfitting, resulting in a noticeable decline in text-image alignment, thereby compromising the model's original generative capabilities. Conversely, differential LoRA training better captures the difference in image pairs and avoids overfitting.
\daoyuan{A natural related question: what if we make some effort to tune hyperparameters, such as decrease the learning rate and steps, the conclusion still hold? As the number of dataset is too small and the tuning cost is also light}

\subsection{Limitations}

This study explores a novel approach to improve text-to-image models, distinct from RLHF and DPO. We aim to replace costly human resources with multimodal LLMs. However, to ensure the controllability of training outcomes, we have retained manual selection and review steps. Due to budget constraints, we employed only two graduate-level computer science students to complete this process, making our approach more lightweight compared to previous studies \cite{black2023training} that required hundreds of individuals for manual annotation. Similar to the pre-training of large models, we believe this process is scalable. By utilizing a larger cluster for data generation and hiring more personnel for selection and review, we could further enhance the model's performance, which remains an avenue for future work.

\begin{table}[]
\caption{Quantitative comparison between differential LoRA training and naive LoRA training.}
\label{tab:differential_training}
\centering
\setlength{\tabcolsep}{3pt}{
\begin{tabular}{l|cccc}
\hline
                       & Aesthetic     & PickScore      & MPS            & CLIP             \\ \hline
Naive training         & 6.25          & 47.15          & 49.81          & 29.34            \\
Differential training  & \textbf{6.35} & \textbf{52.85} & \textbf{50.19} & \textbf{29.71}   \\ \hline
\end{tabular}}
\end{table}

\section{Conclusion}

In this paper, we explore a method to enhance text-to-image models. To guide these models in generating high-quality images that align with human preferences, we introduce \ours. By employing a synthesis-understanding interaction algorithm, we improve the image quality, albeit at the expense of increased inference time. This interaction algorithm enables the creation of a dataset specifically for training, thereby facilitating a differential training approach to learn the enhancement. By iterating this process, we progressively refine the text-to-image model. Based on state-of-the-art text-to-image models, we trained \ours modules in the form of LoRA. Experimental results highlight the substantial improvements achieved through \ours. Our approach effectively attains alignment training with minimal human resource costs.

\nocite{langley00}

\bibliography{example_paper}
\bibliographystyle{icml2024}

\newpage
\appendix
\onecolumn

\section{Examples of interactions}
\label{app:examples}

\begin{figure*}[p]
    \centering
    \begin{tabular}{m{0.2cm}<{\centering} m{0.29\textwidth}<{\centering} m{0.29\textwidth}<{\centering} m{0.29\textwidth}<{\centering}}
        \rotatebox{90}{Before interaction} & 
        \includegraphics[width=0.29\textwidth]{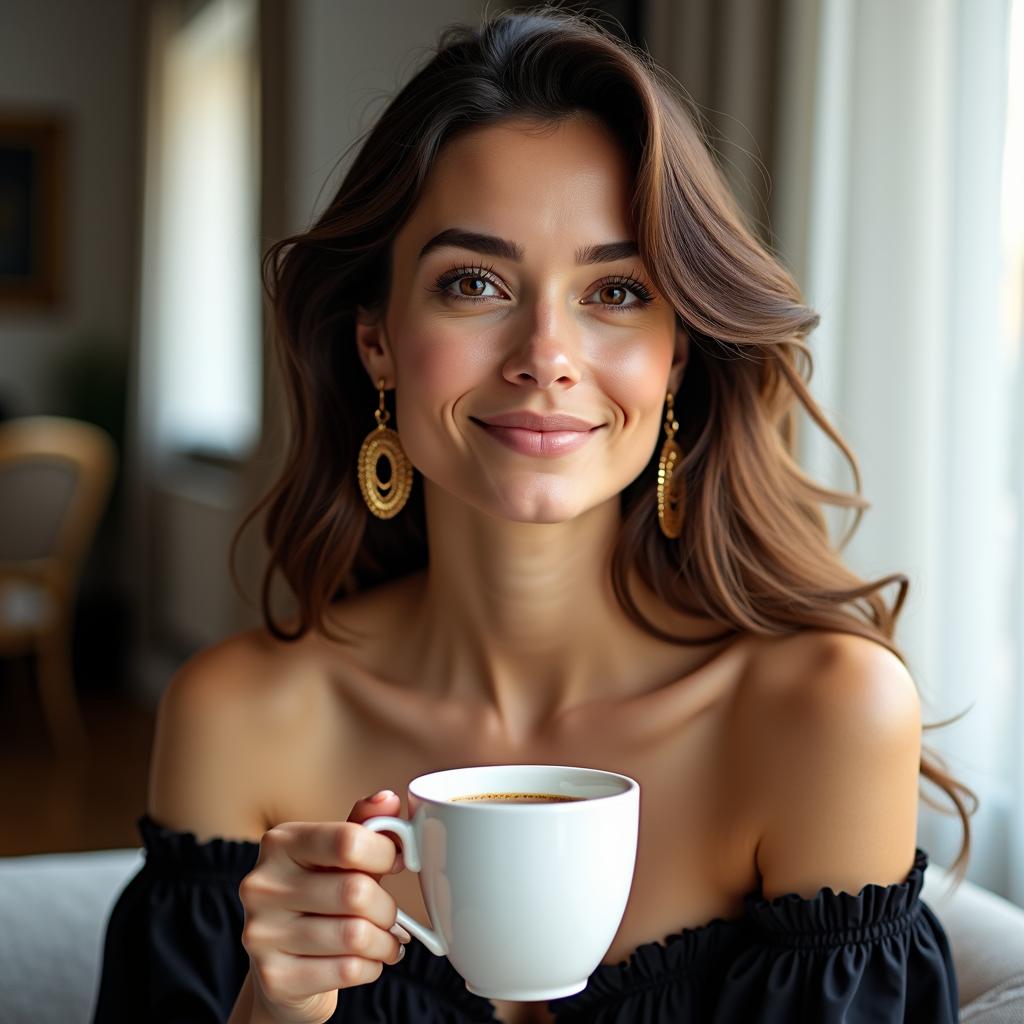} & 
        \includegraphics[width=0.29\textwidth]{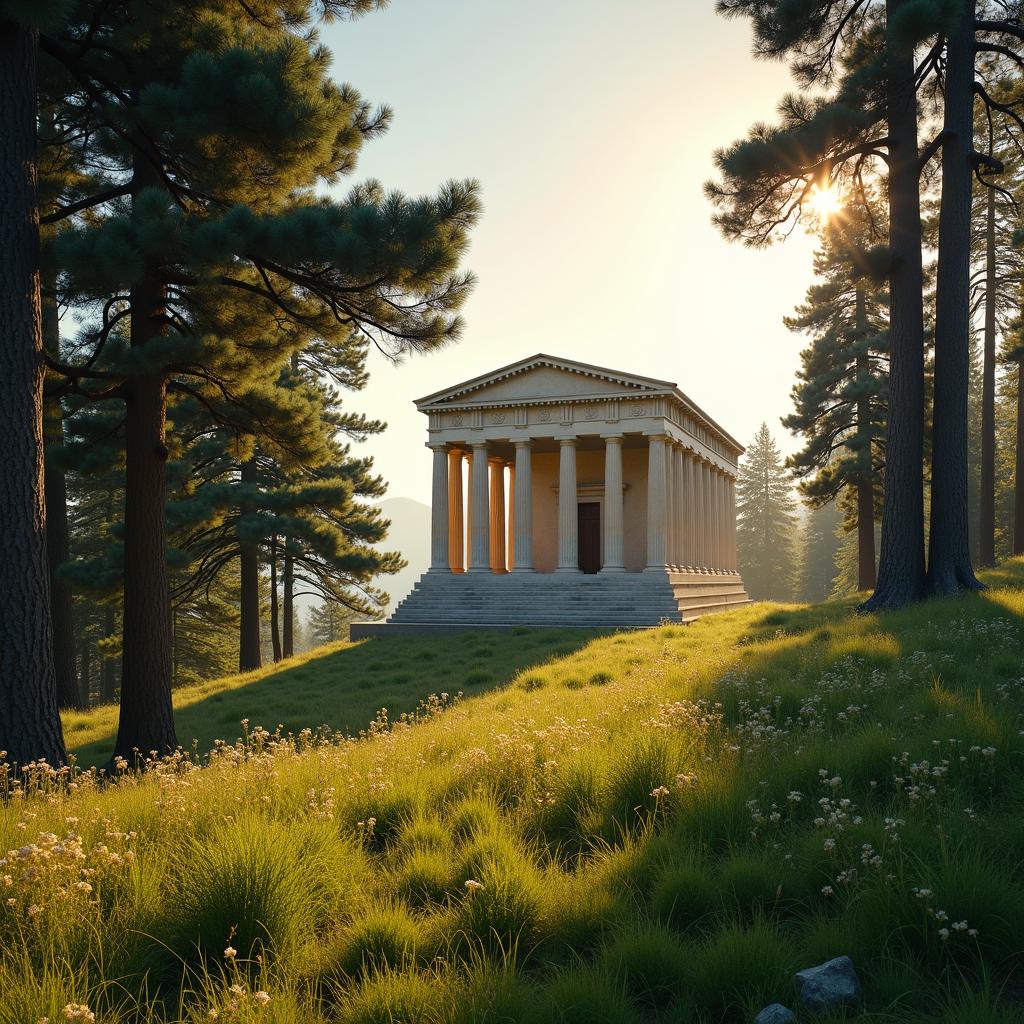} & 
        \includegraphics[width=0.29\textwidth]{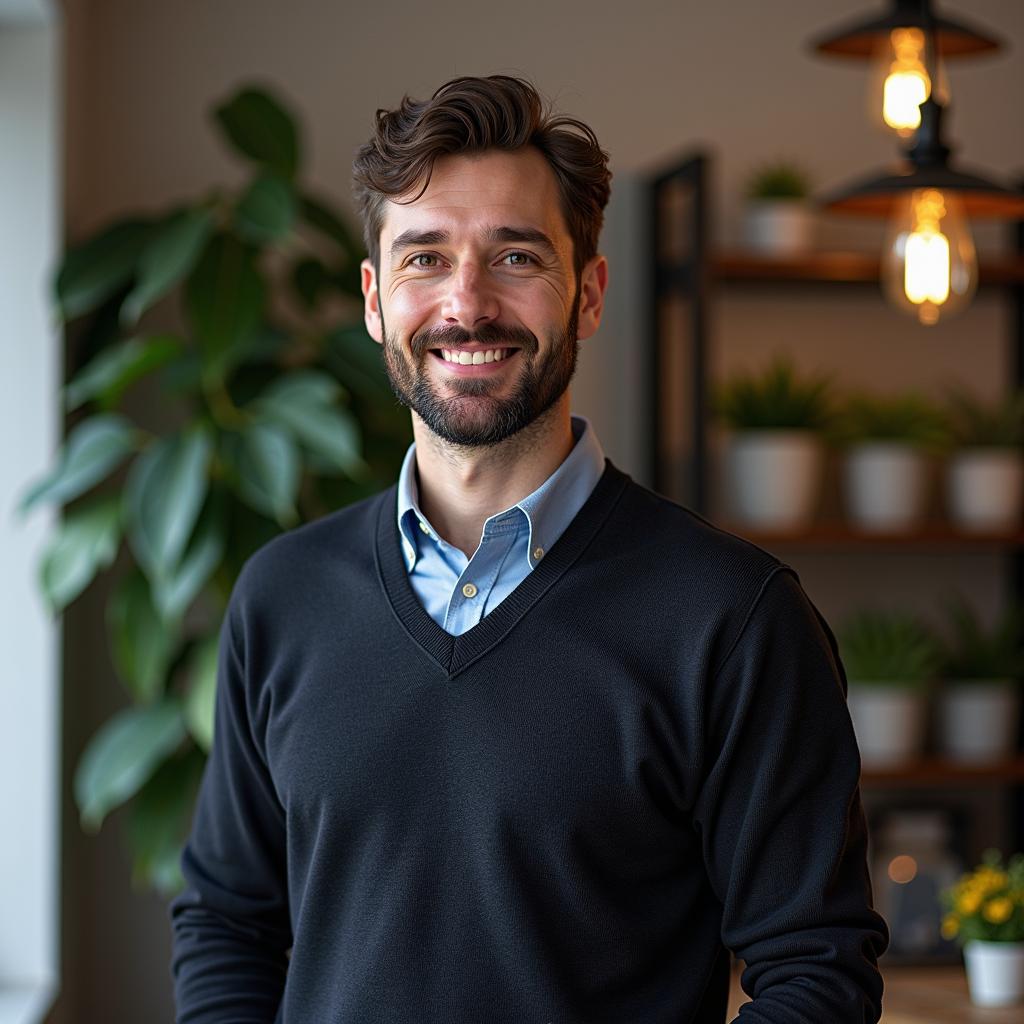} \\
        \vspace{0.2cm}
        
        \rotatebox{90}{After interaction} & 
        \includegraphics[width=0.29\textwidth]{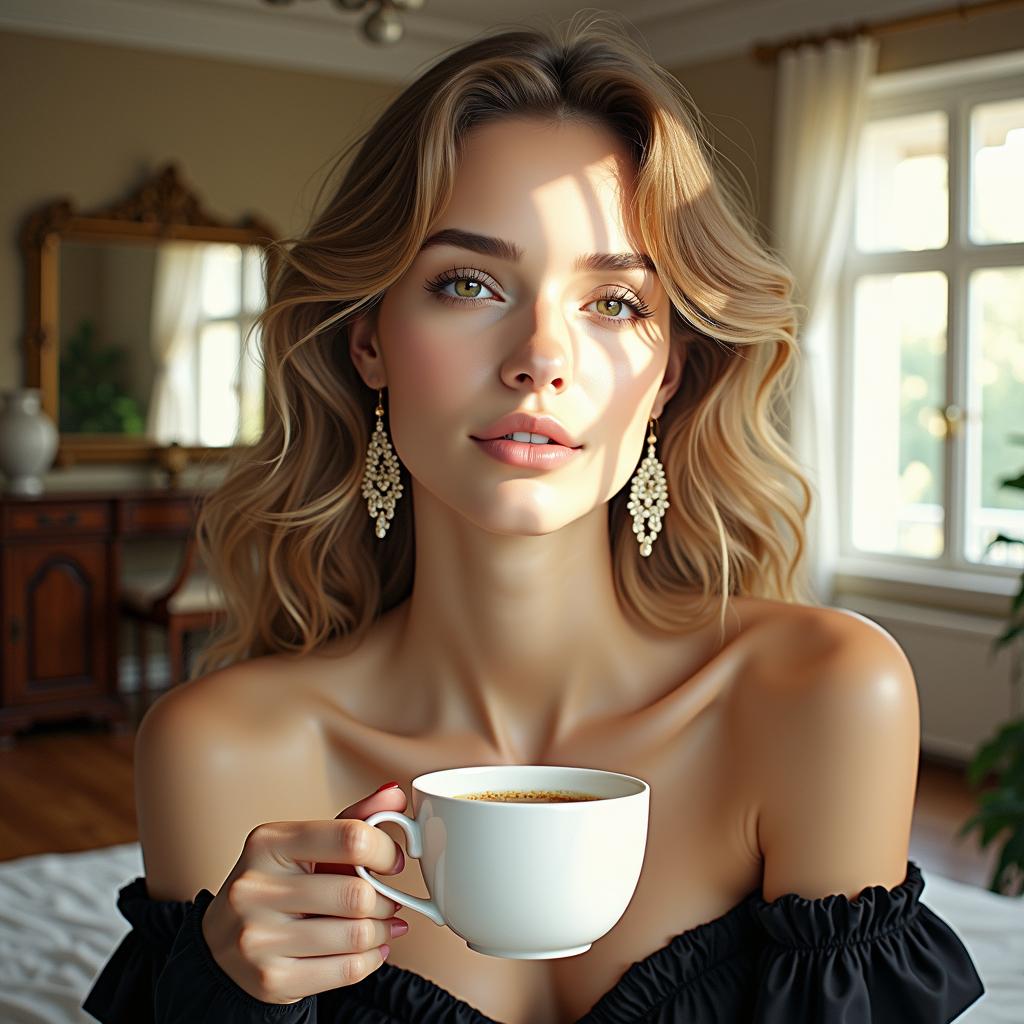} & 
        \includegraphics[width=0.29\textwidth]{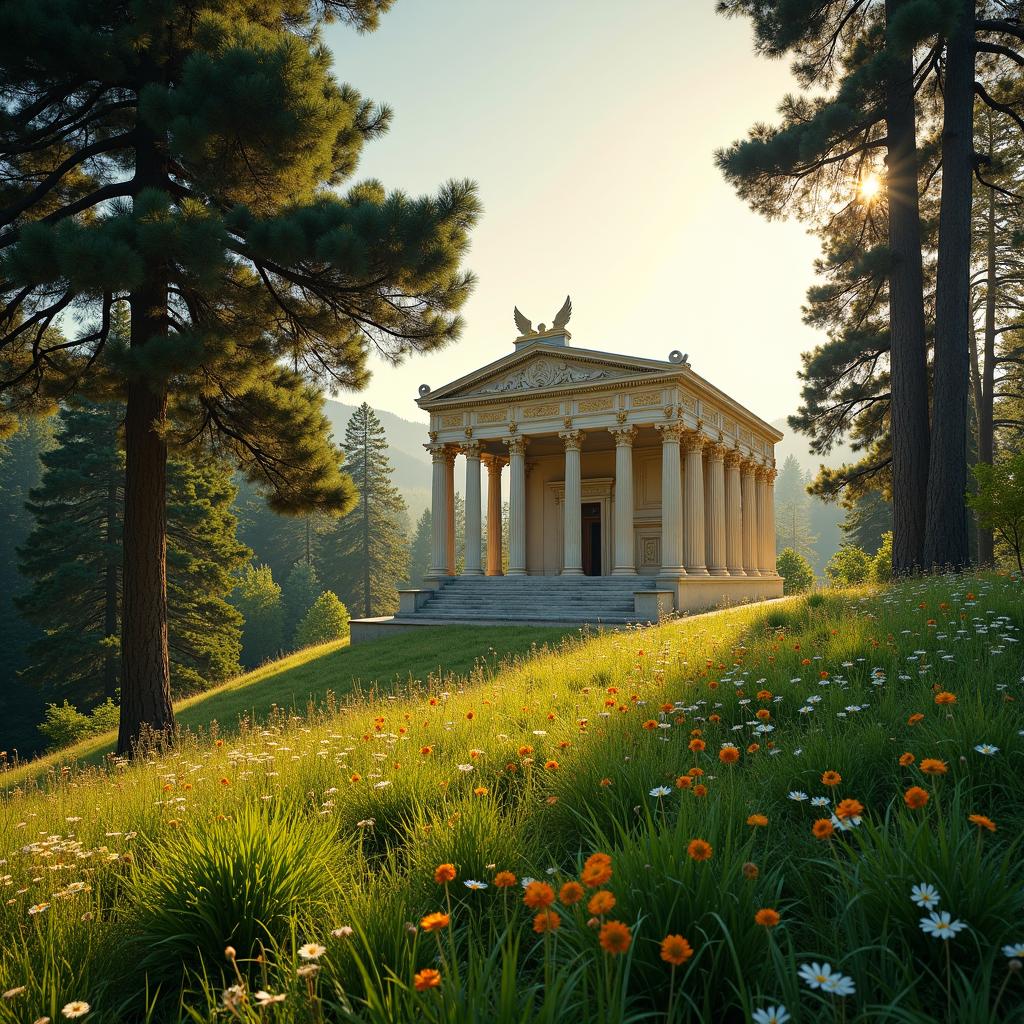} & 
        \includegraphics[width=0.29\textwidth]{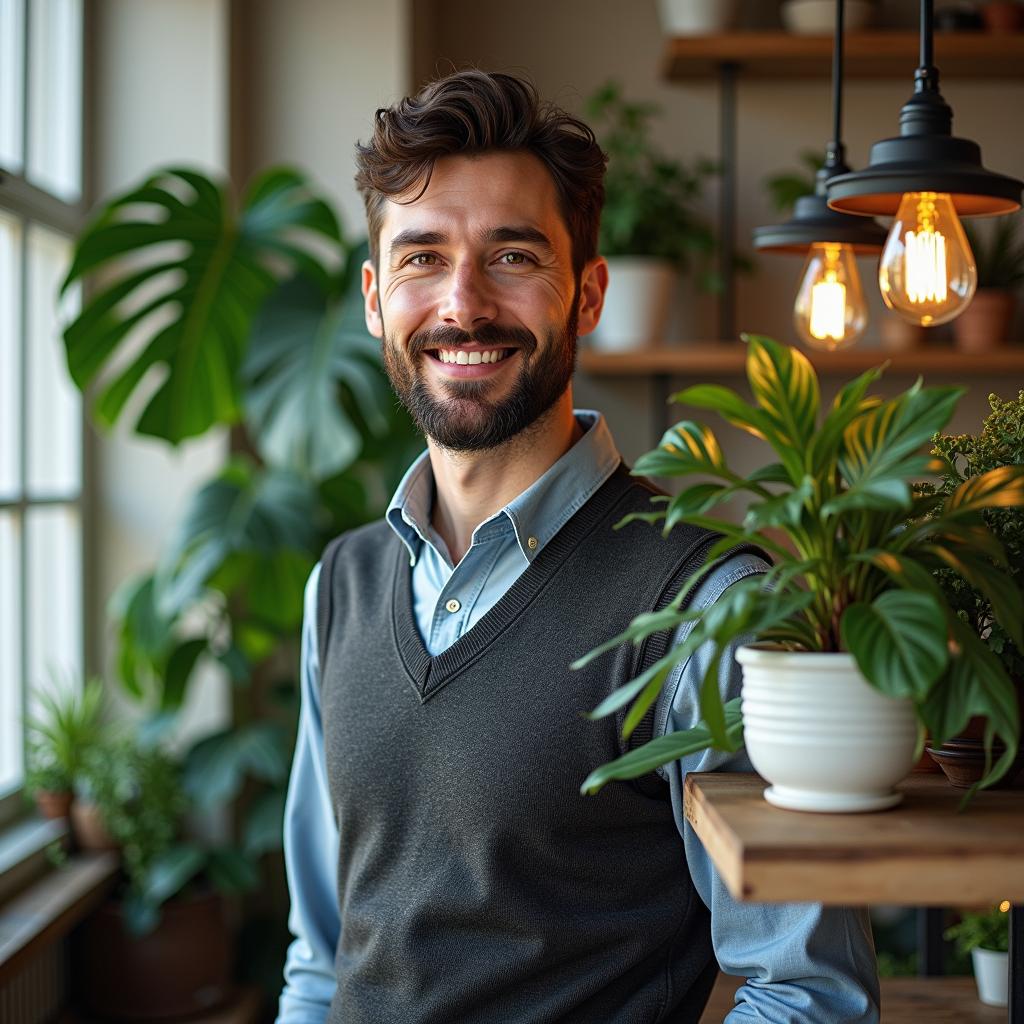} \\
        
        & (a) Lighting & (b) Detail & (c) Composition \\
        
        \rotatebox{90}{Before interaction} & 
        \includegraphics[width=0.29\textwidth]{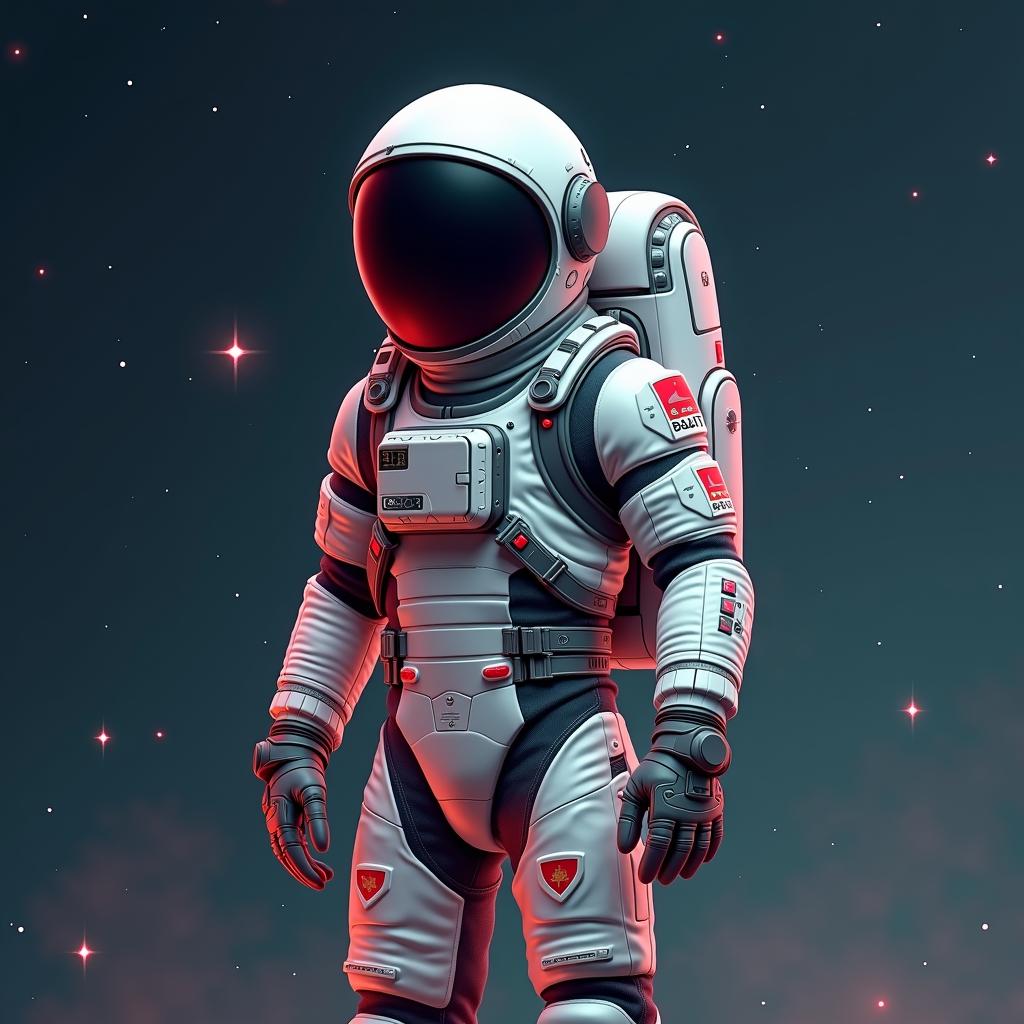} & 
        \includegraphics[width=0.29\textwidth]{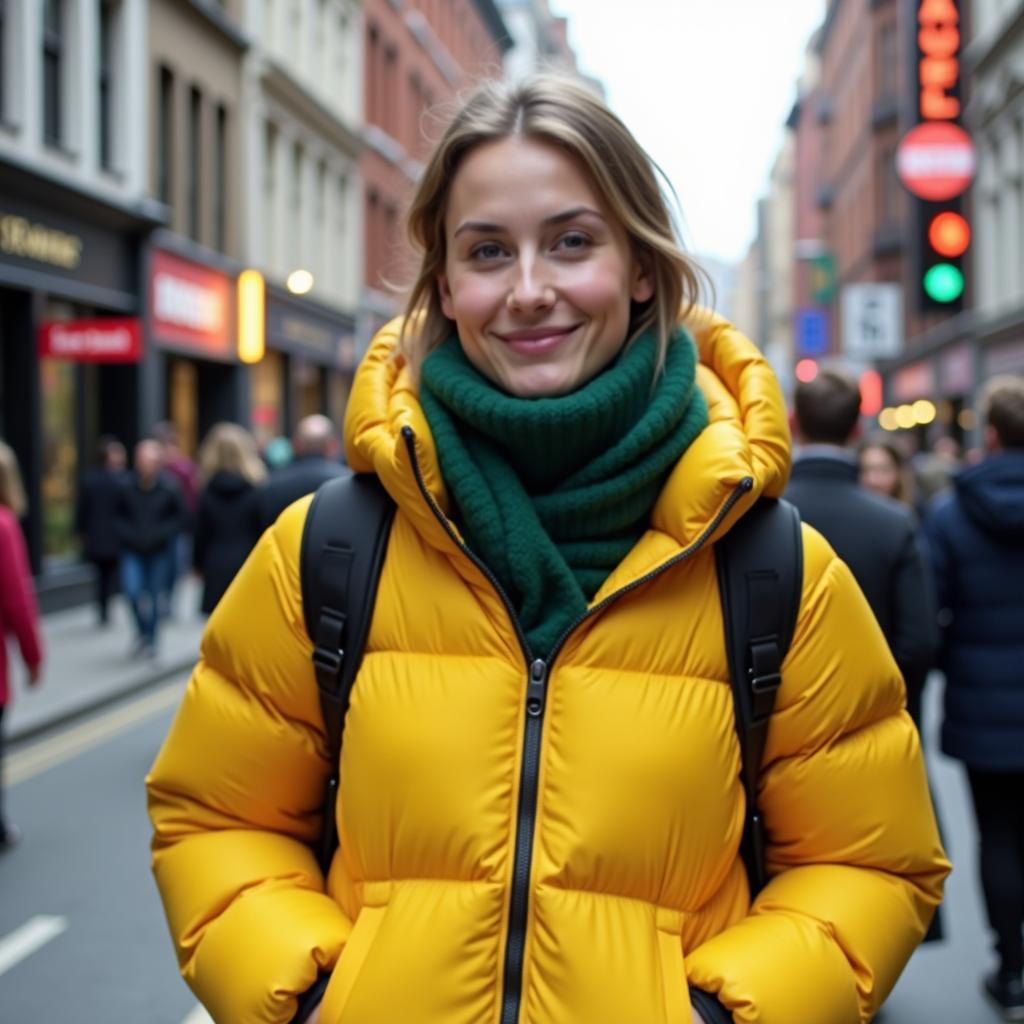} & 
        \includegraphics[width=0.29\textwidth]{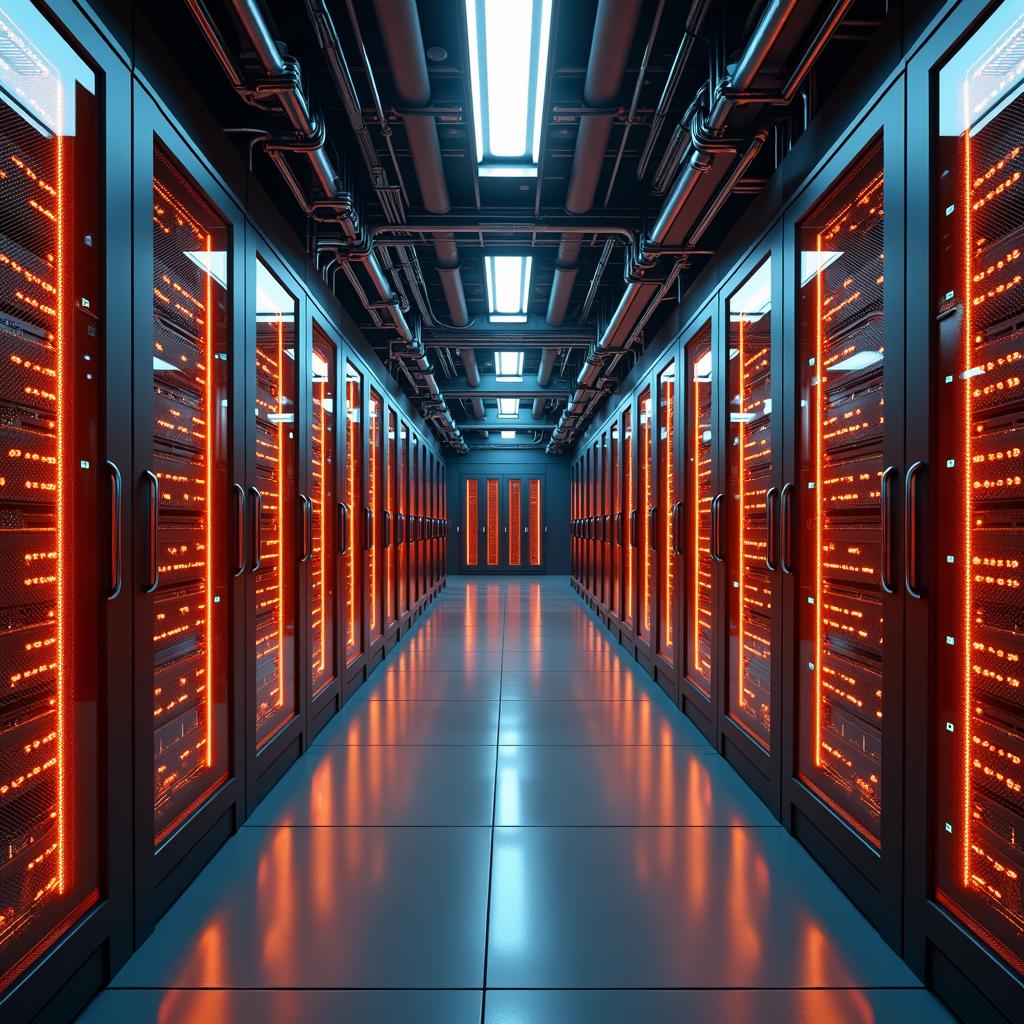} \\
        \vspace{0.2cm}
        
        \rotatebox{90}{After interaction} & 
        \includegraphics[width=0.29\textwidth]{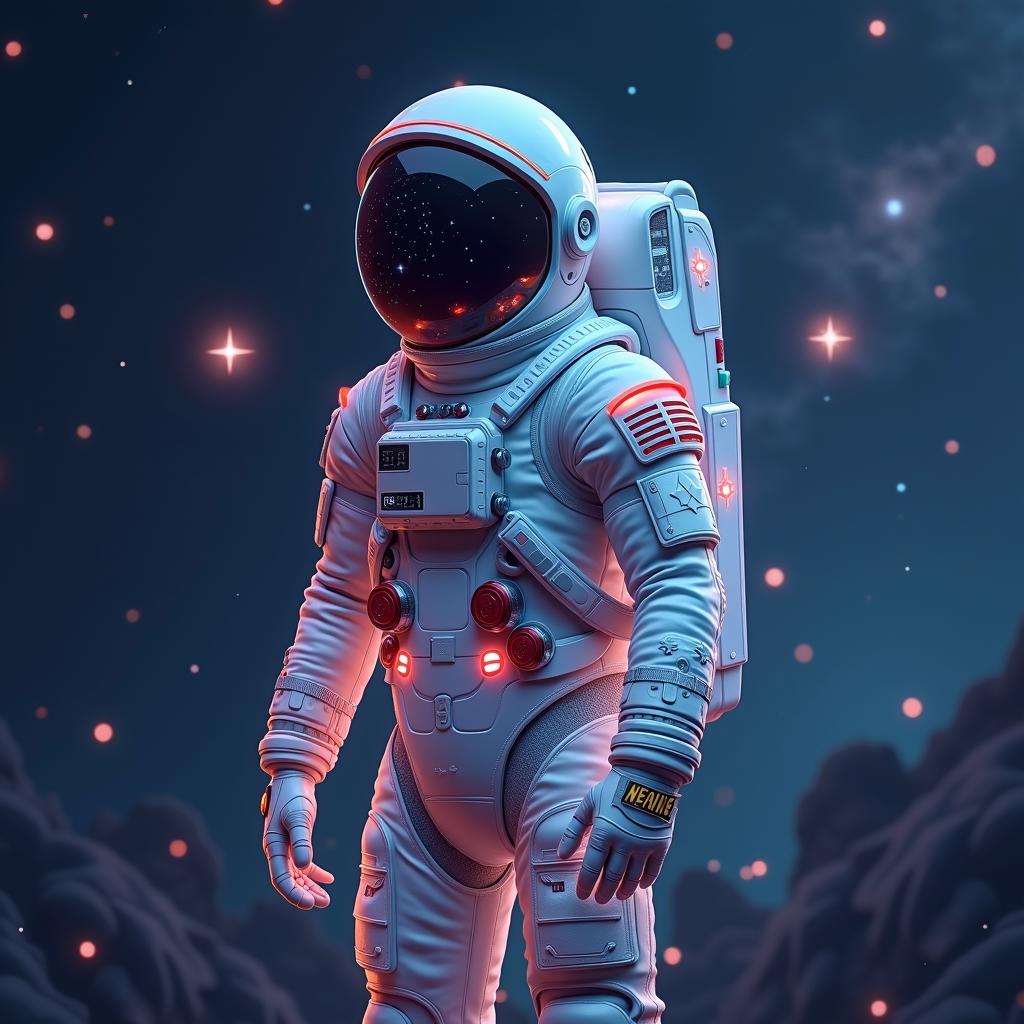} & 
        \includegraphics[width=0.29\textwidth]{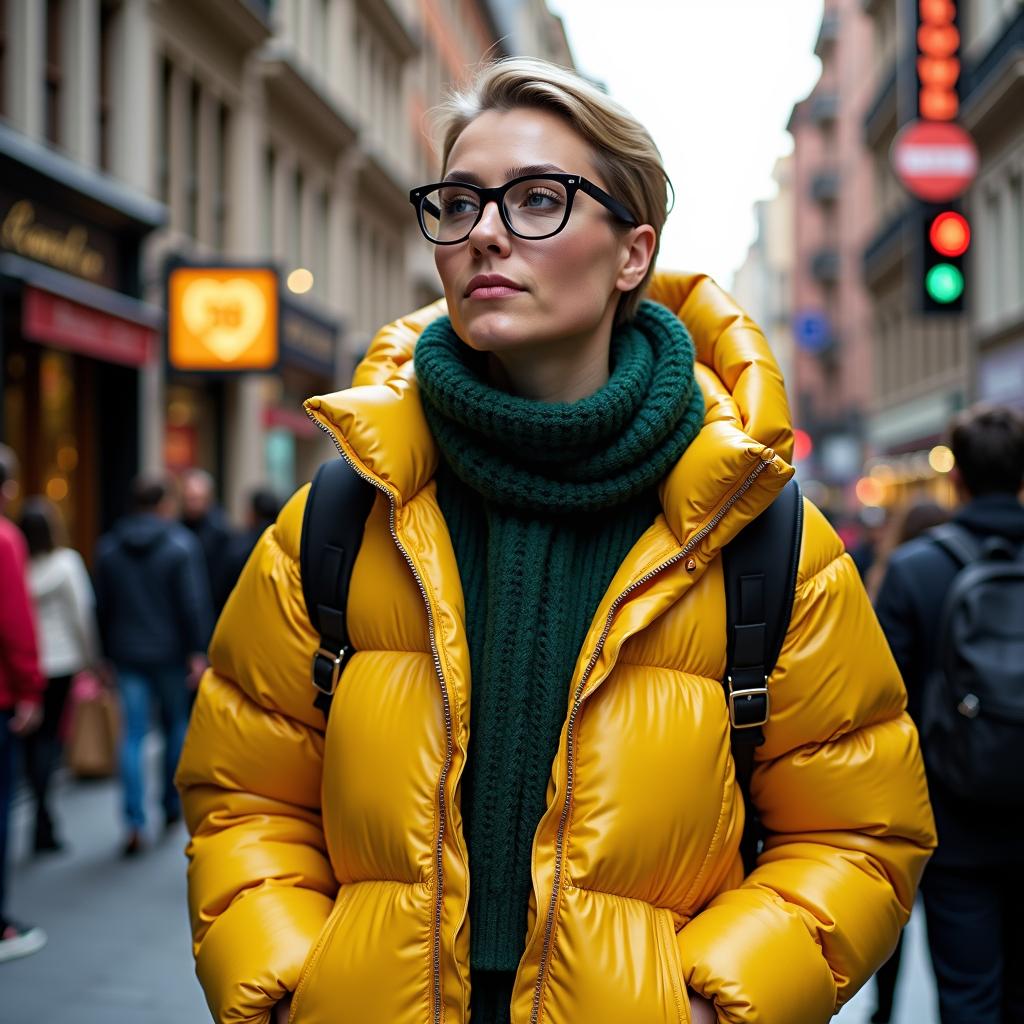} & 
        \includegraphics[width=0.29\textwidth]{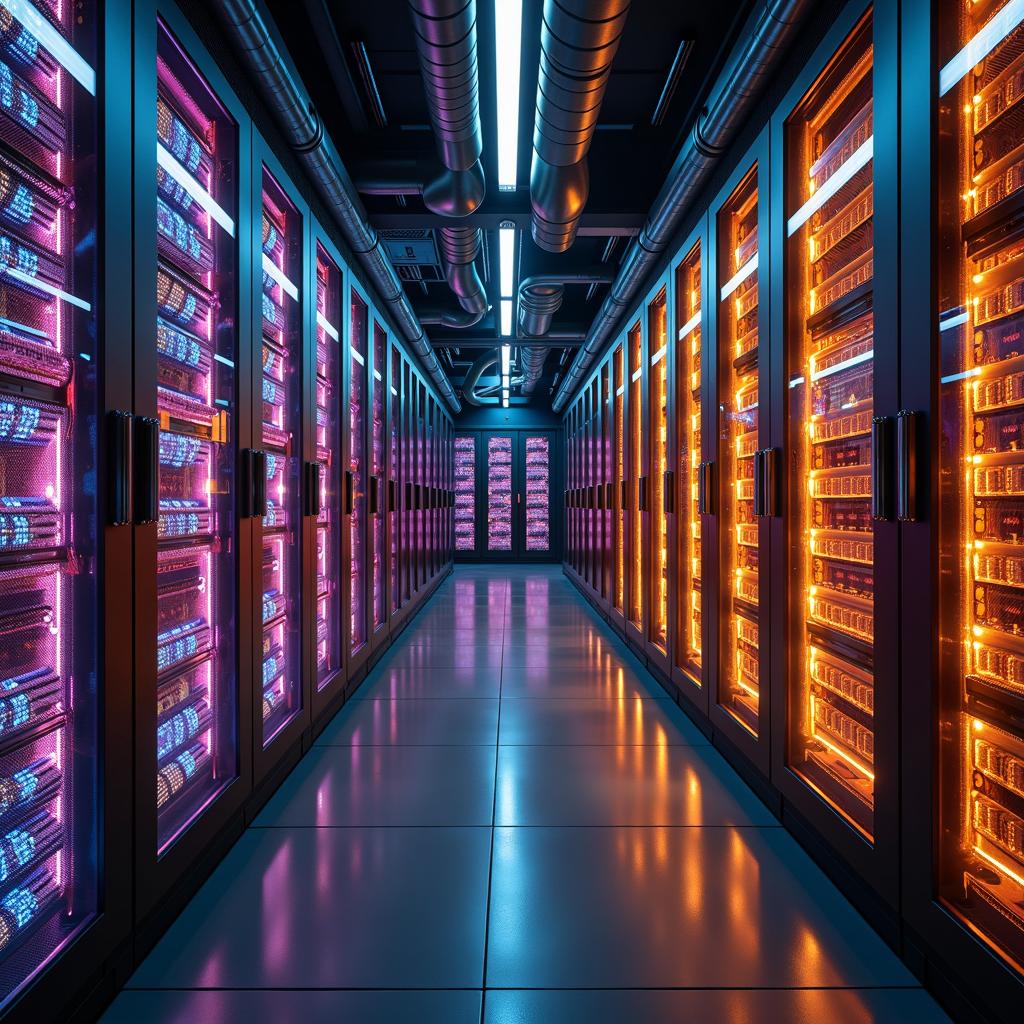} \\

        & (d) Ambiance & (e) Clarity & (f) Color \\
    \end{tabular}
    \caption{Image examples improved by the interaction algorithm. The enhanced images exhibit aesthetic improvements in various aspects.}
    \label{fig:examples1}
\end{figure*}
\begin{figure*}[p]
    \centering
    \begin{tabular}{m{0.2cm}<{\centering} m{0.29\textwidth}<{\centering} m{0.29\textwidth}<{\centering} m{0.29\textwidth}<{\centering}}
        \rotatebox{90}{Before interaction} & 
        \includegraphics[width=0.29\textwidth]{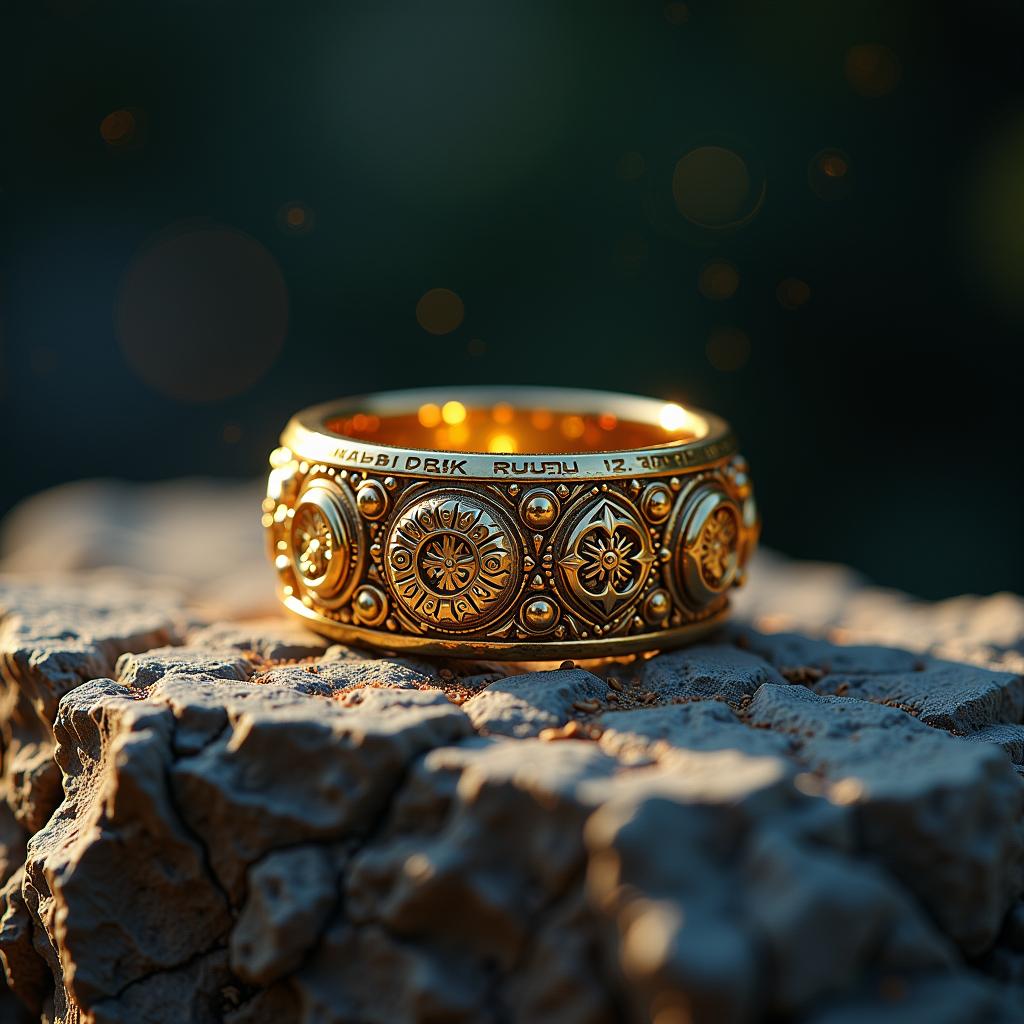} & 
        \includegraphics[width=0.29\textwidth]{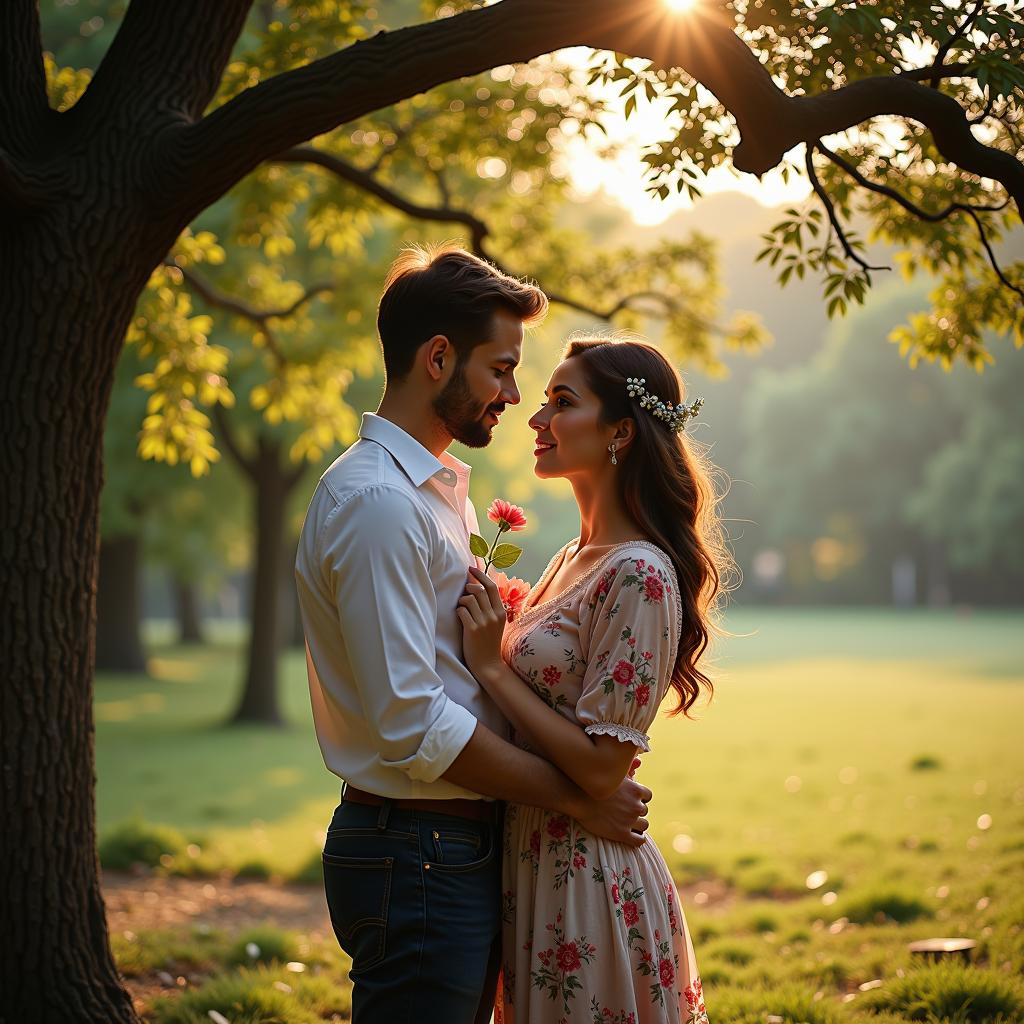} & 
        \includegraphics[width=0.29\textwidth]{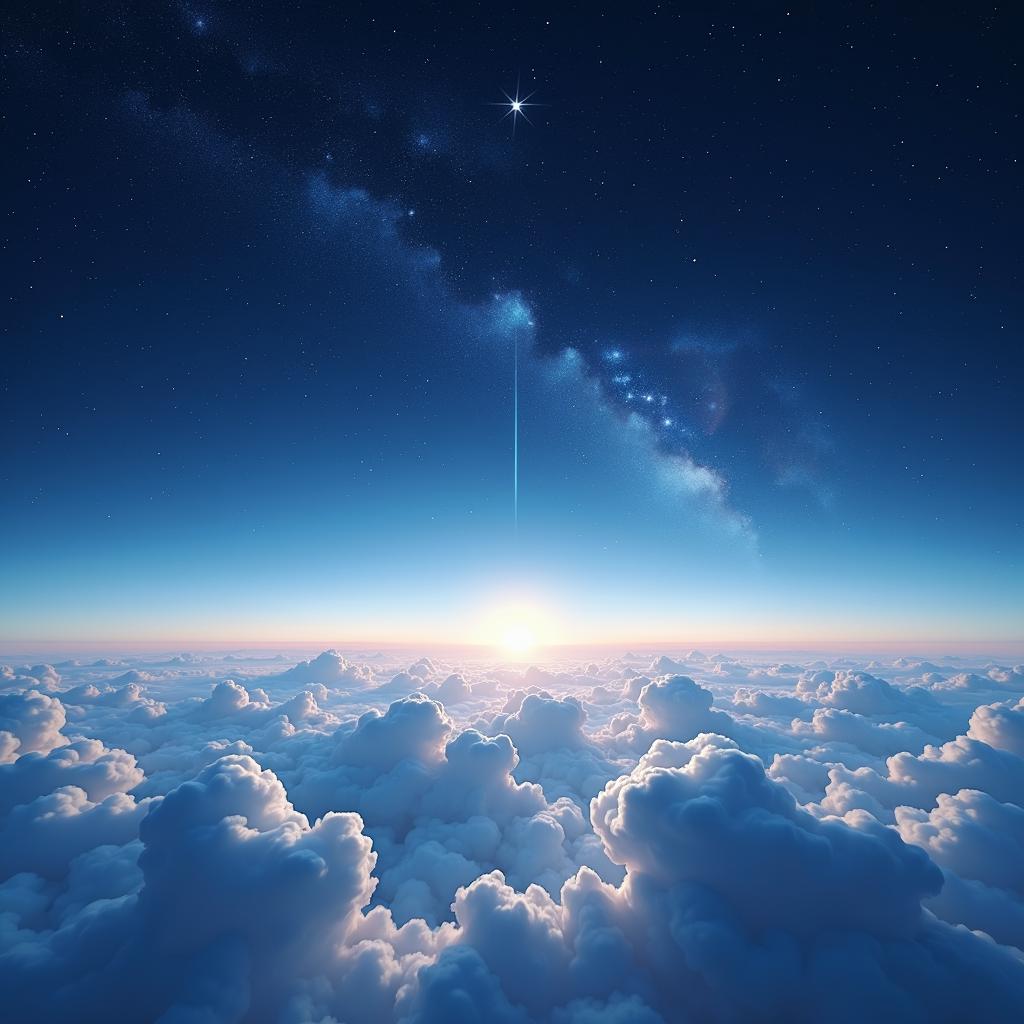} \\
        \vspace{0.2cm}
        
        \rotatebox{90}{After interaction} & 
        \includegraphics[width=0.29\textwidth]{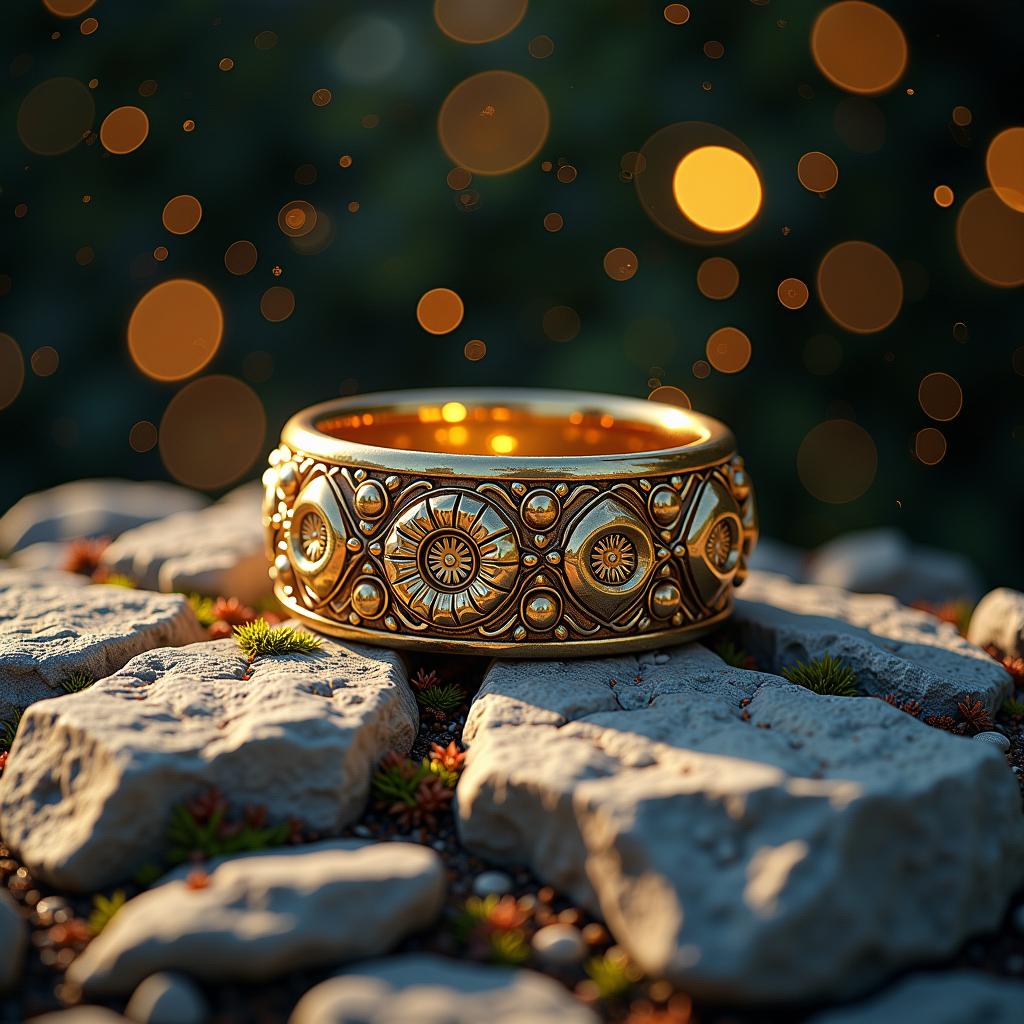} & 
        \includegraphics[width=0.29\textwidth]{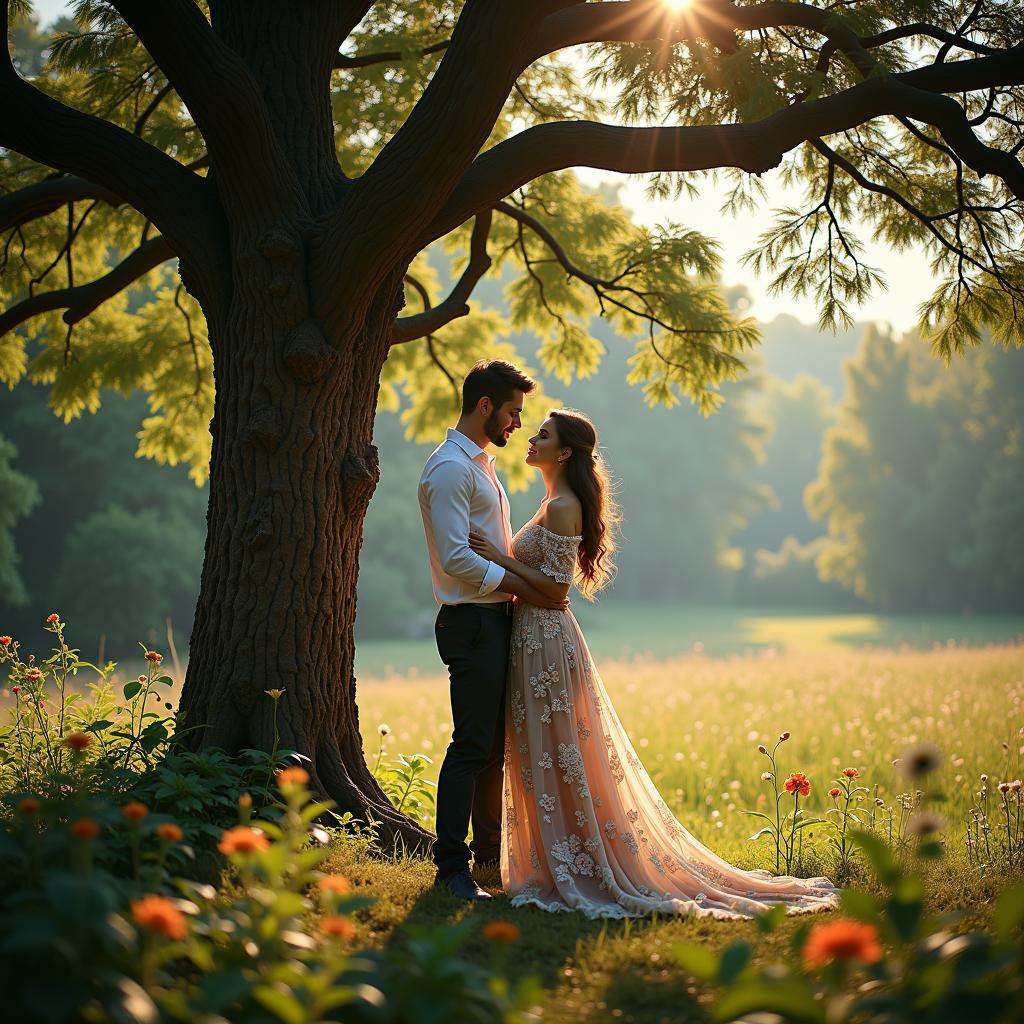} & 
        \includegraphics[width=0.29\textwidth]{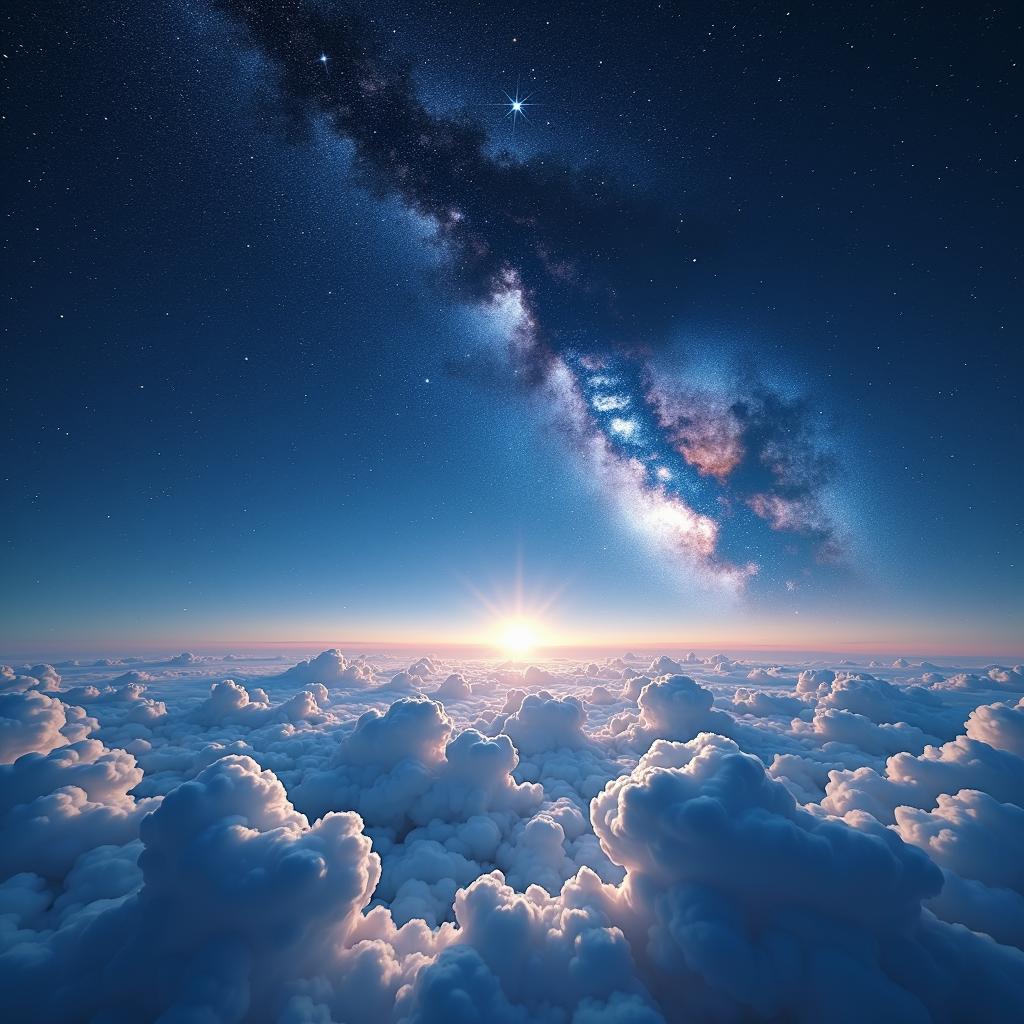} \\
        
        & (a) Particle effects & (b) Shooting angle & (c) Exposure compensation \\
        
        \rotatebox{90}{Before interaction} & 
        \includegraphics[width=0.29\textwidth]{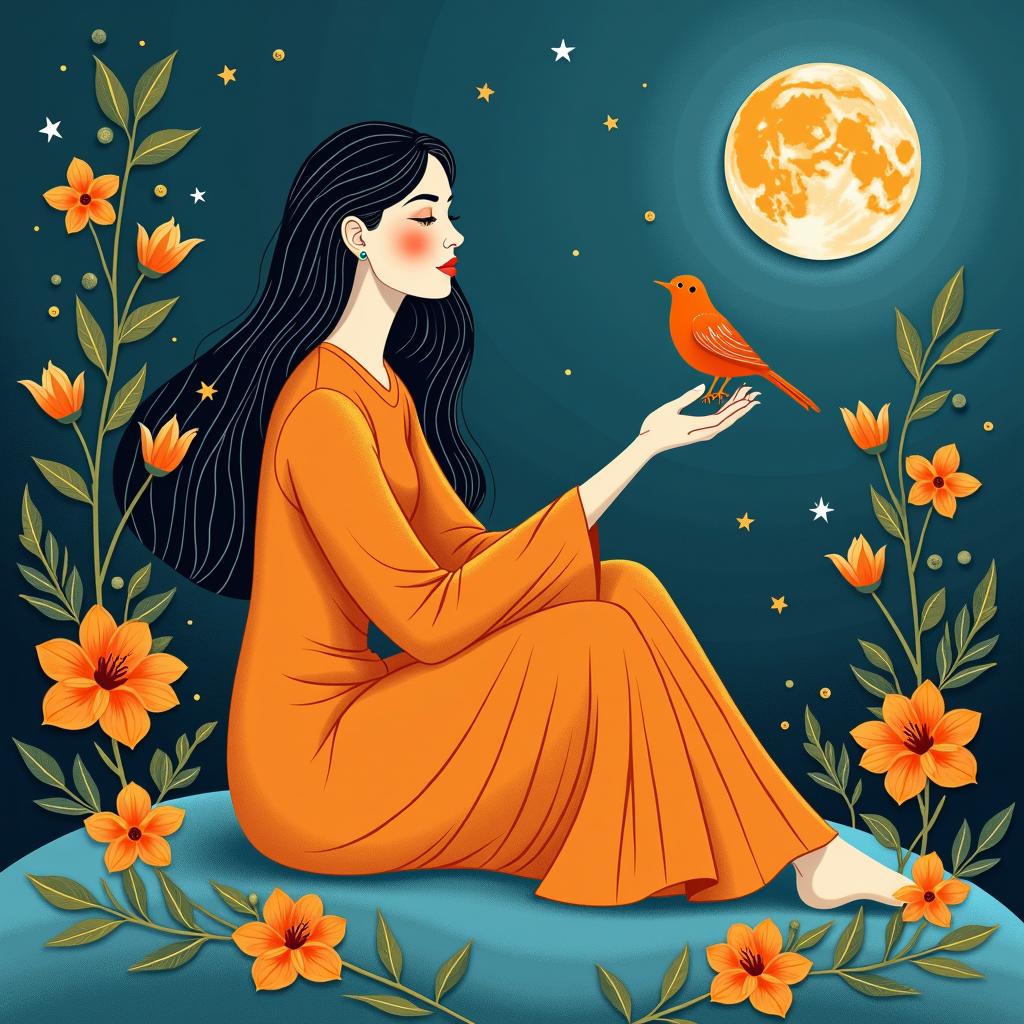} & 
        \includegraphics[width=0.29\textwidth]{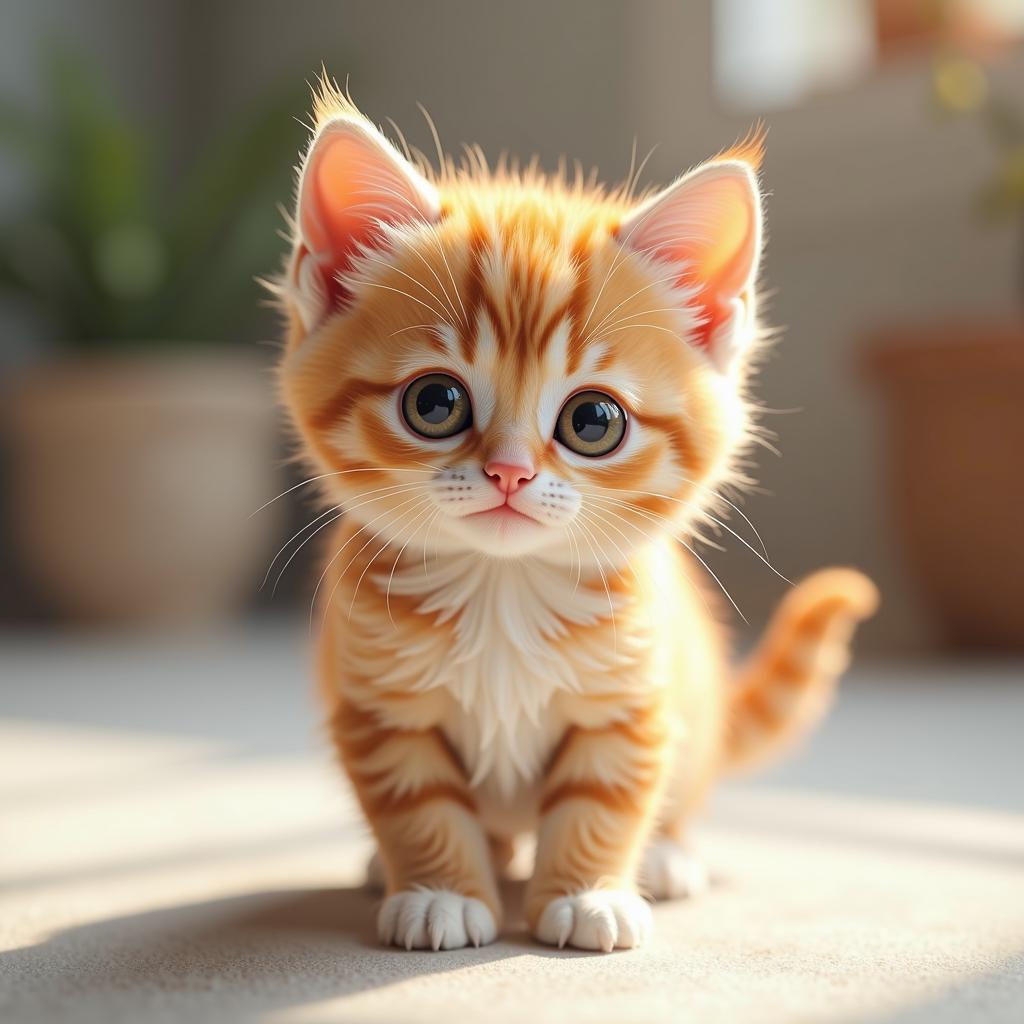} & 
        \includegraphics[width=0.29\textwidth]{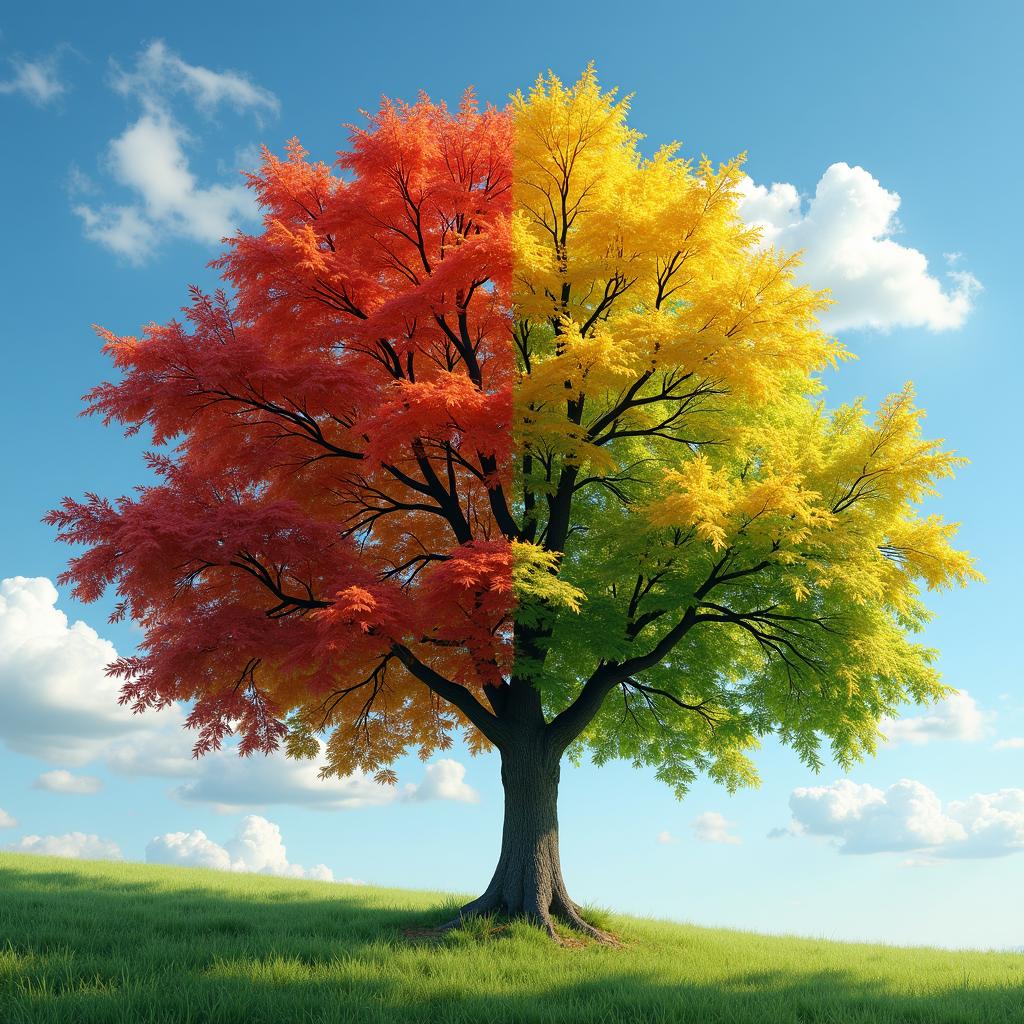} \\
        \vspace{0.2cm}
        
        \rotatebox{90}{After interaction} & 
        \includegraphics[width=0.29\textwidth]{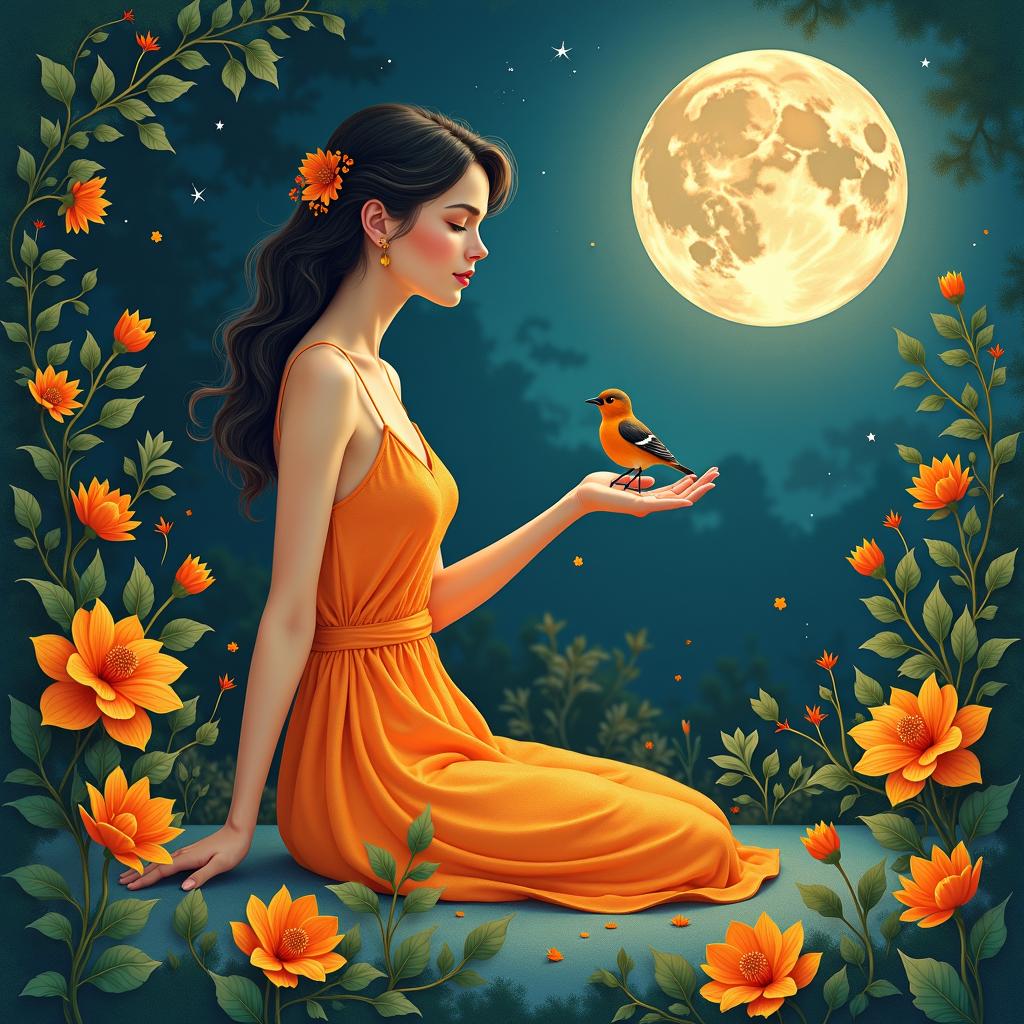} & 
        \includegraphics[width=0.29\textwidth]{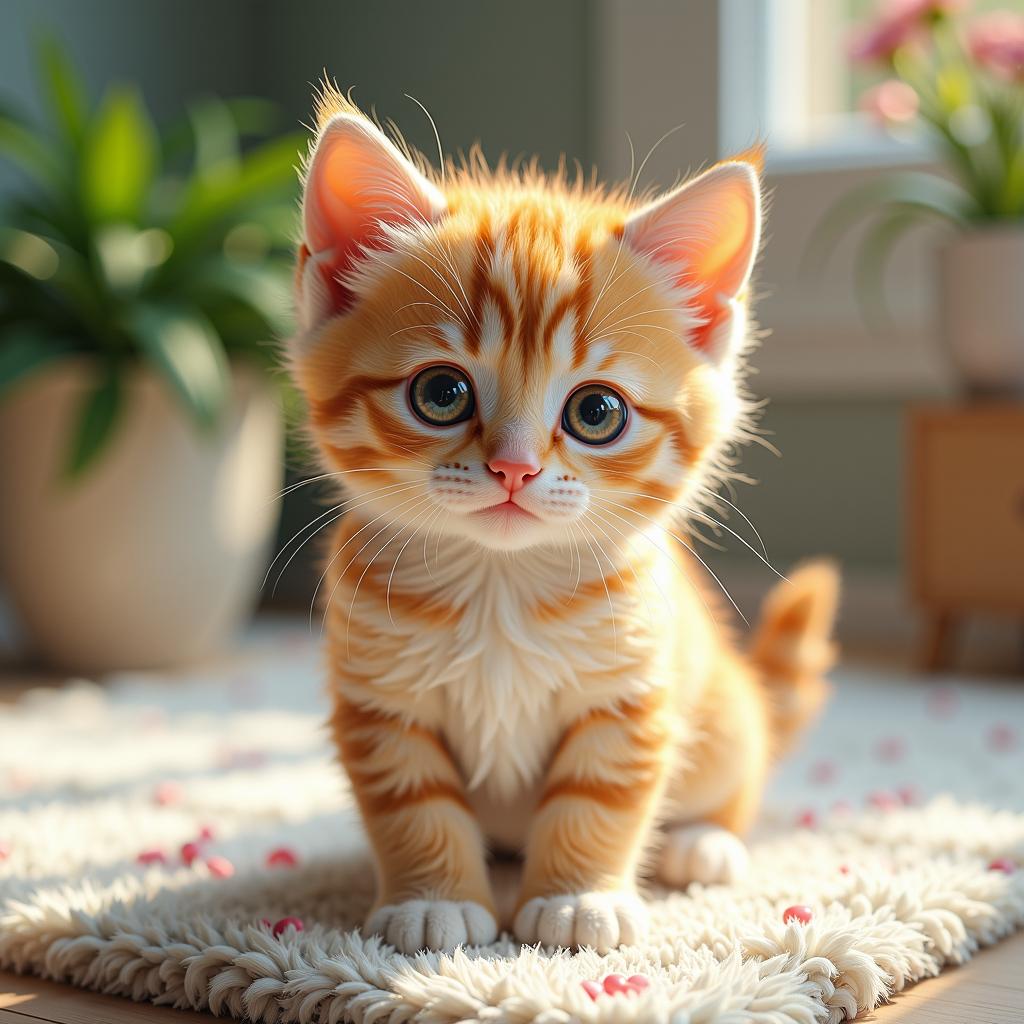} & 
        \includegraphics[width=0.29\textwidth]{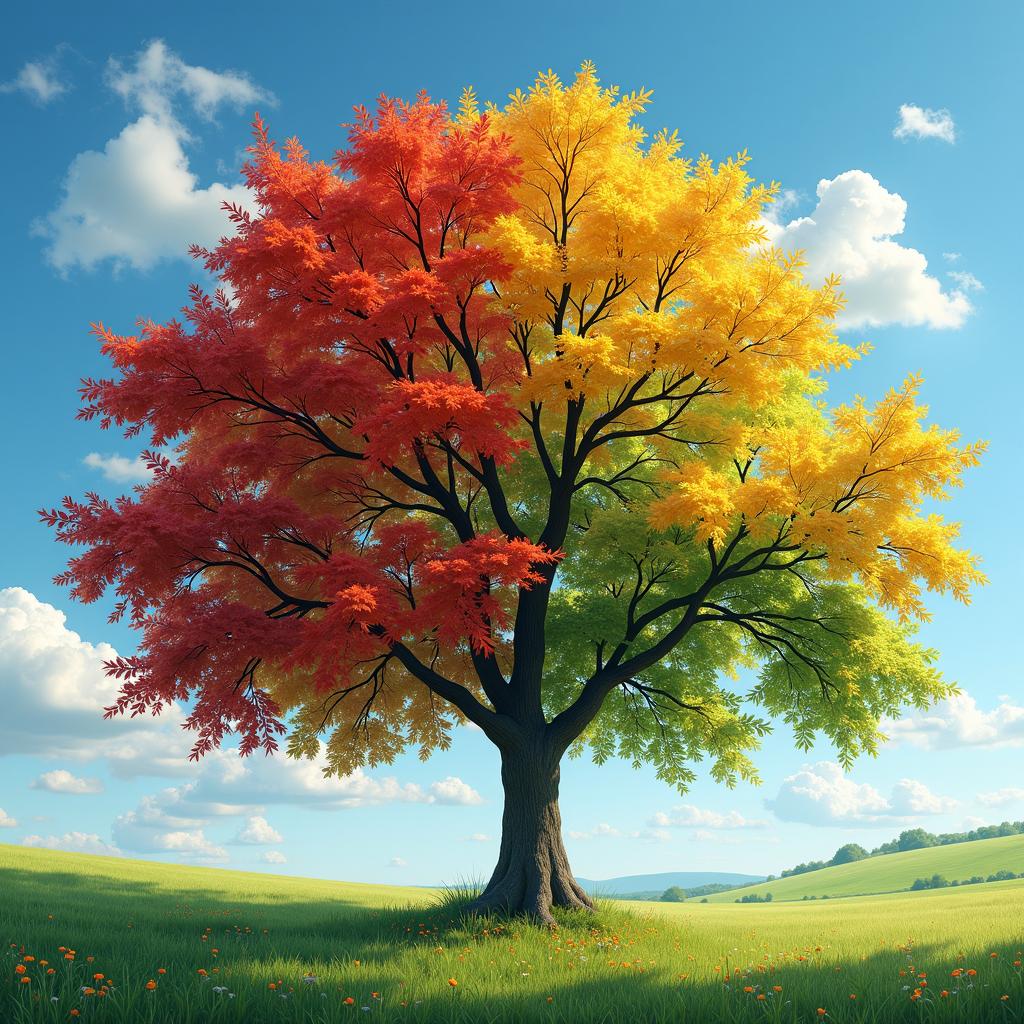} \\

        & (d) Style adjustment & (e) Background blur & (f) Color gradient \\
    \end{tabular}
    \caption{Image examples improved by the interaction algorithm. The multimodal LLM in the interaction algorithm is capable of understanding human preferences and making fine-grained adjustments to the content of images.}
    \label{fig:examples2}
\end{figure*}

Some image pairs generated by our proposed interaction algorithm are displayed in Figures \ref{fig:examples1} and \ref{fig:examples2}. These images clearly demonstrate that the multimodal LLM can enhance image quality across various fundamental aesthetic aspects. The improvements in the basic aesthetic aspects include:
\begin{itemize}
\item \textbf{Lighting}: Optimizes the effects of natural and artificial light, ensuring a balance of highlights and shadows.

\item \textbf{Detail}: Enhances subtle yet crucial elements of objects in the image, boosting realism and visual appeal.

\item \textbf{Composition}: Adjusts the relative positions of objects within the image, enhancing compositional effects and achieving balanced spatial aesthetics.

\item \textbf{Ambiance}: Optimizes the background and atmosphere of the image, creating an environment and mood that matches the theme.

\item \textbf{Clarity}: Improves overall clarity, reducing noise and blur.

\item \textbf{Color}: Adjusts temperature, saturation, and more, resulting in vibrant, harmonious colors while retaining the original scene's atmosphere.
\end{itemize}
Beyond these fundamental aesthetic enhancements, our algorithm achieves more advanced effects, including but not limited to:
\begin{itemize}
\item \textbf{Particle Effects}: Introduces dynamic or special effects, such as particle effects, to images.

\item \textbf{Shooting Angle}: Alters camera angles for a richer visual experience.

\item \textbf{Exposure Compensation}: Simulates realistic scenarios like a galaxy appearing with increased exposure.

\item \textbf{Style Adjustment}: Converts images to specific artistic styles to make them aesthetically pleasing.

\item \textbf{Background Blur}: Highlights main subjects while ensuring natural transitions in the background.

\item \textbf{Color Gradient}: Employs color gradients to smoothly transition between colors, resulting in a softer and more harmonious image.
\end{itemize}

These improvements highlight how well multimodal LLMs can enhance image aesthetics and adapt content and style to suit human preferences. The interactive algorithm effectively transfers the multimodal LLMs' understanding of aesthetics to the text-to-image model, thereby guiding the image generation process.

\section{Prompt of Multimodal LLMs}
\label{app:prompt}

The prompt used in Qwen2-VL-72B has undergone several iterations and extensive testing to ensure its effectiveness in guiding the model to generate enriched and aesthetically pleasing details in the image. This prompt is detailed as follows, where ``\_\_prompt\_\_'' denotes the original prompt of the text-to-image model. 

\begin{lstlisting}
You are a helpful assistant. Given the image please analyze the following image and complete the following tasks:

1. Add more details to this image. For example, beautiful light and shadow, exquisite decorations, gorgeous clothing, beautiful natural landscapes, etc. Caution:
    The added details should be consistent with the original description: __prompt__
2. Mark the locations where these details can be added. Caution:
    Each entity should have only a bounding box in the format [x1, y1, x2, y2] represented using absolute pixel coordinates.
3. For each bounding box, imagine that we modify it into something extremely aesthetically pleasing. Please describe the image content of this part using words. Do not use 'should'. Just describe it. The aesthetical description should be long.

Please provide the results in JSON format as follows, which can be directly loaded by json.loads() in Python:
[
    {
        "bbox": [x1, y1, x2, y2],
        "aesthetical description": "..."
    },
    {
        "bbox": [x1, y1, x2, y2],
        "aesthetical description": "..."
    },
    ...
]
\end{lstlisting}

\section{Comparison of Different Multimodal LLMs}
\label{app:mllm}

\begin{figure*}[]
    \centering
    \setlength{\tabcolsep}{3pt}
    \begin{tabular}{ccc}

        \includegraphics[width=0.31\linewidth]{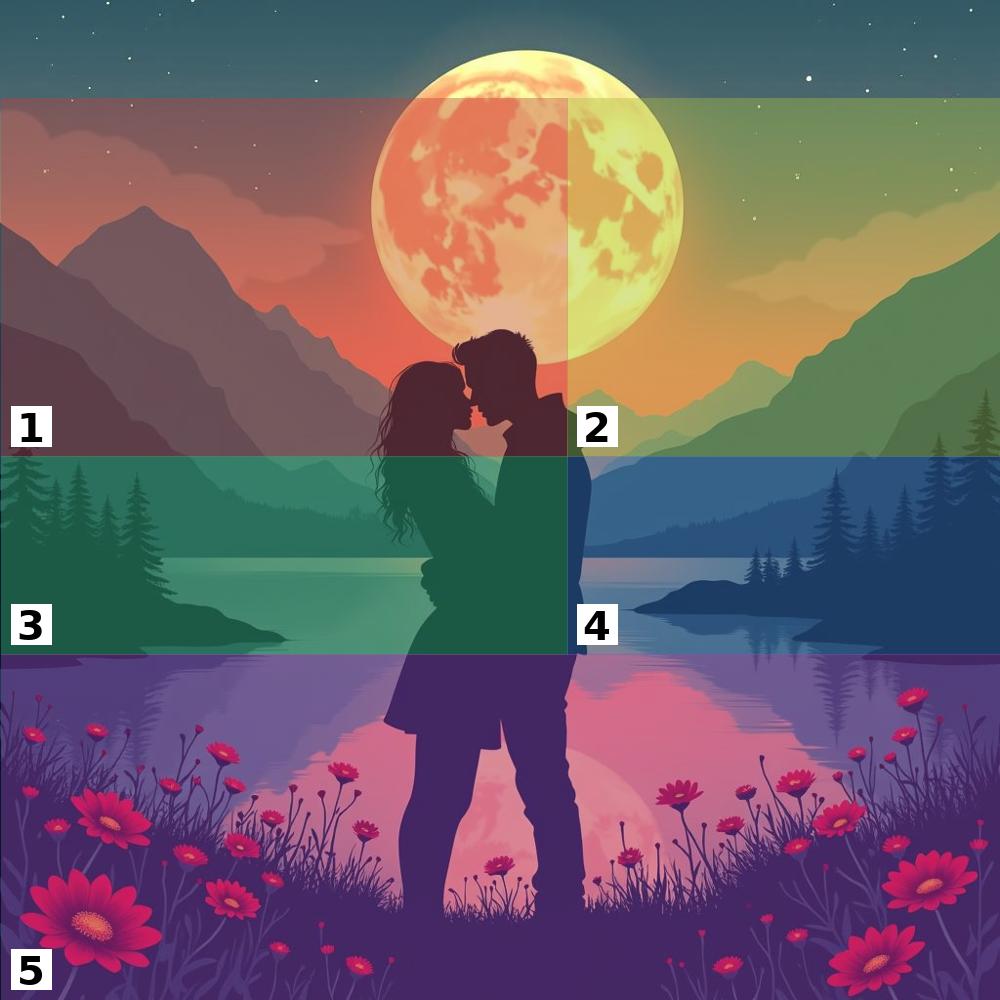} & 
        \includegraphics[width=0.31\linewidth]{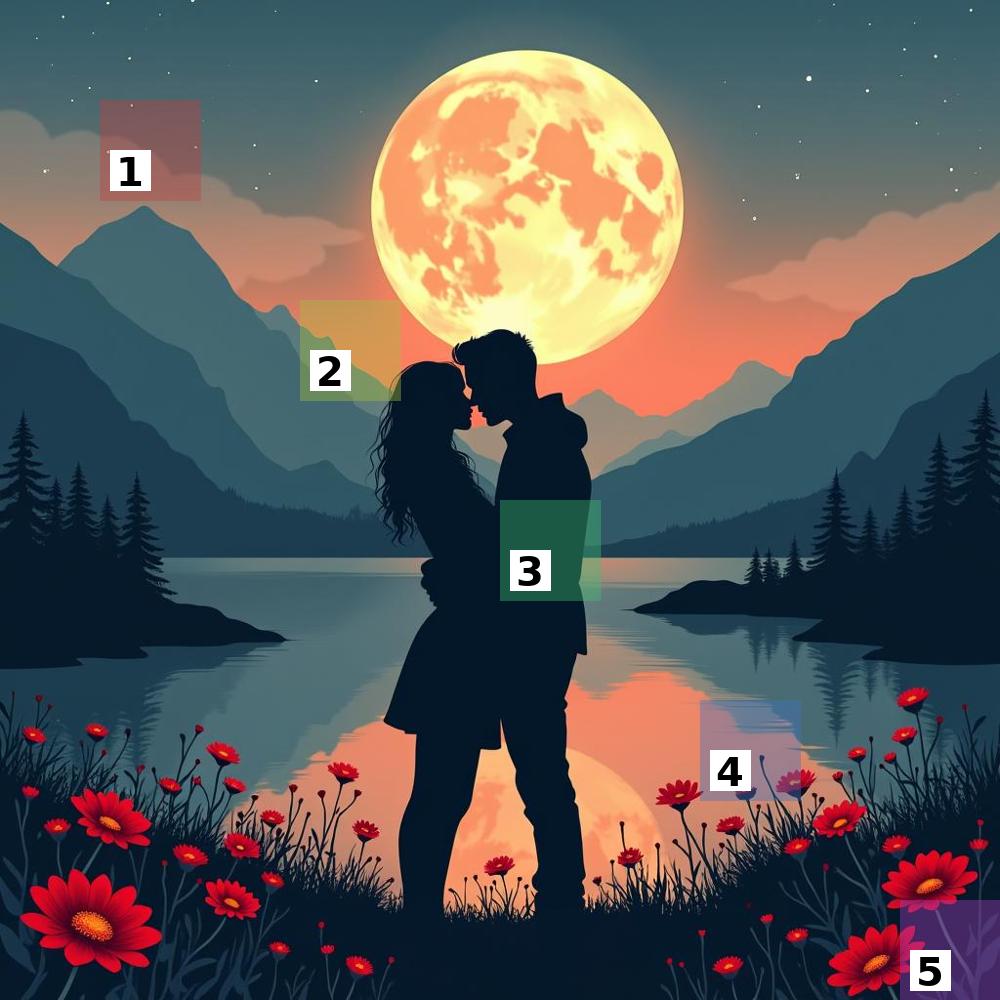} &
        \includegraphics[width=0.31\linewidth]{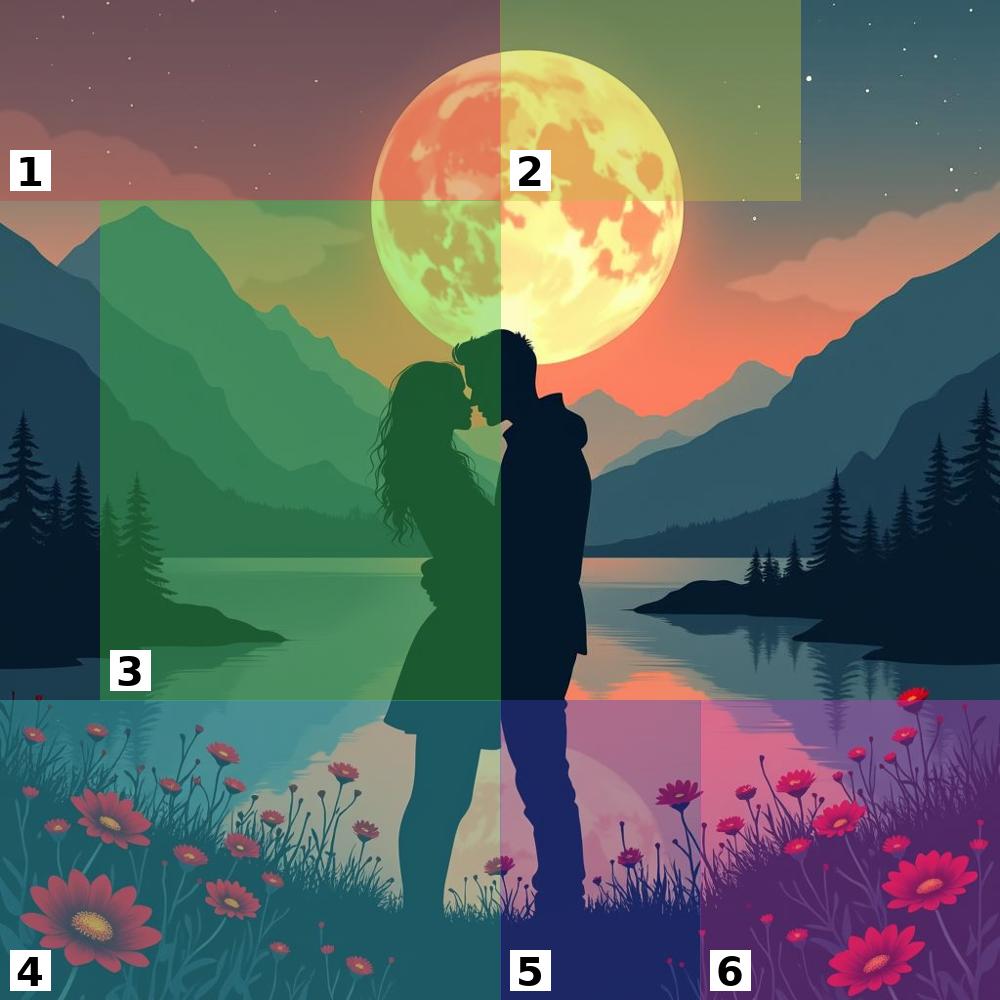} \\

        (a) Qwen2-VL-7B & (b) LLaMa3.2-vision-90b-instruct & (c) InternVL2-26B \\

        \includegraphics[width=0.31\linewidth]{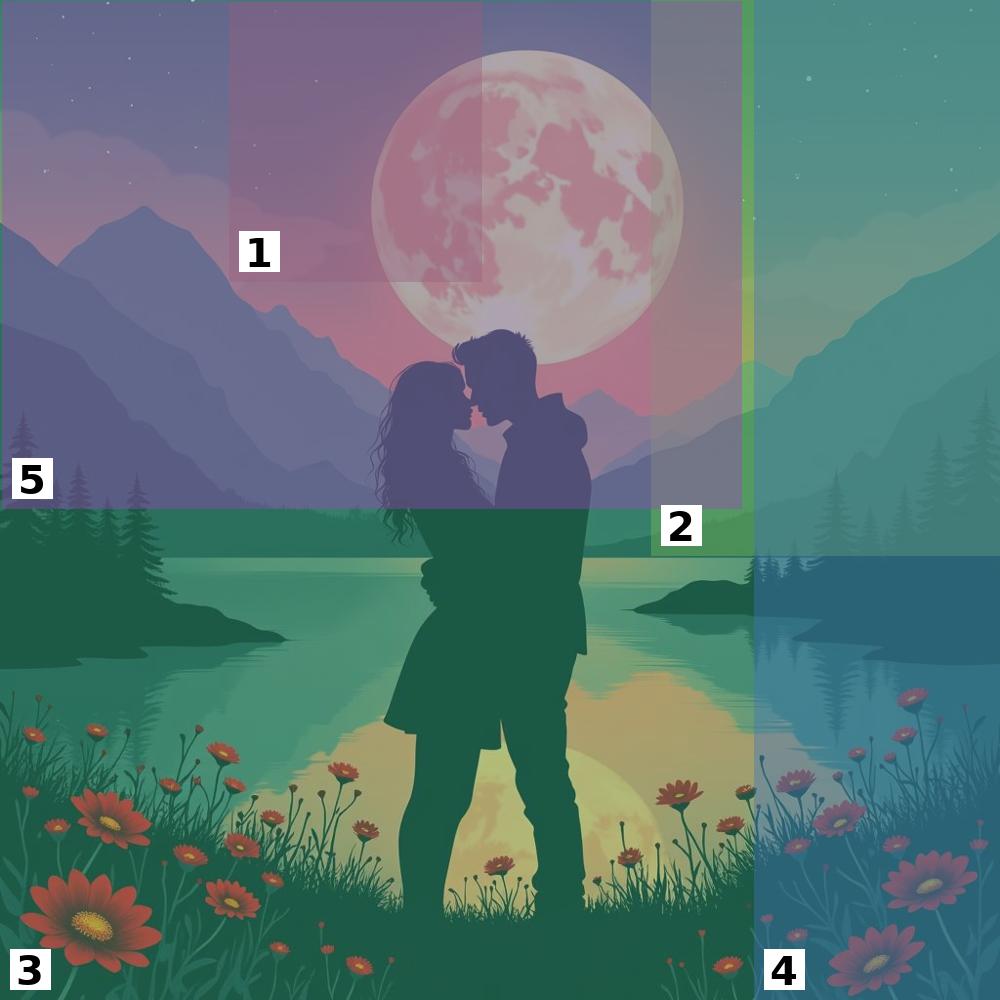} & 
        \includegraphics[width=0.31\linewidth]{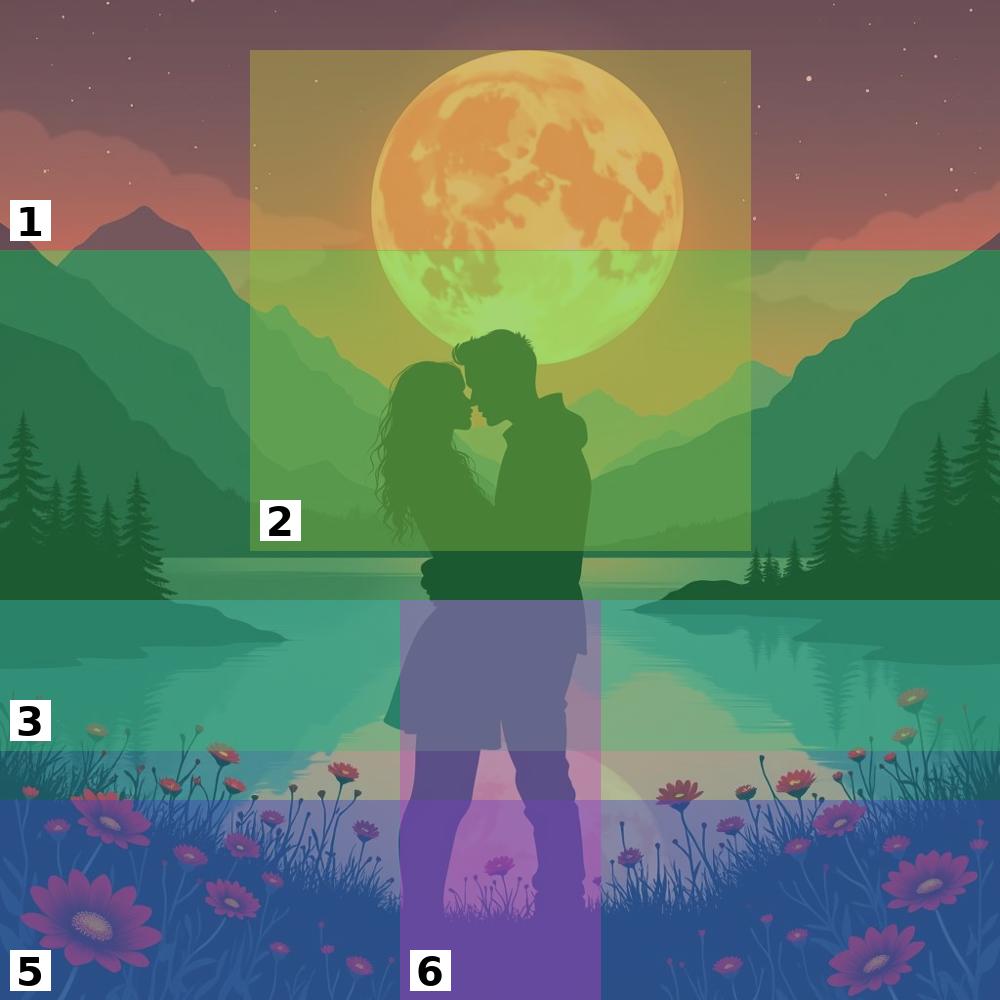} &
        \includegraphics[width=0.31\linewidth]{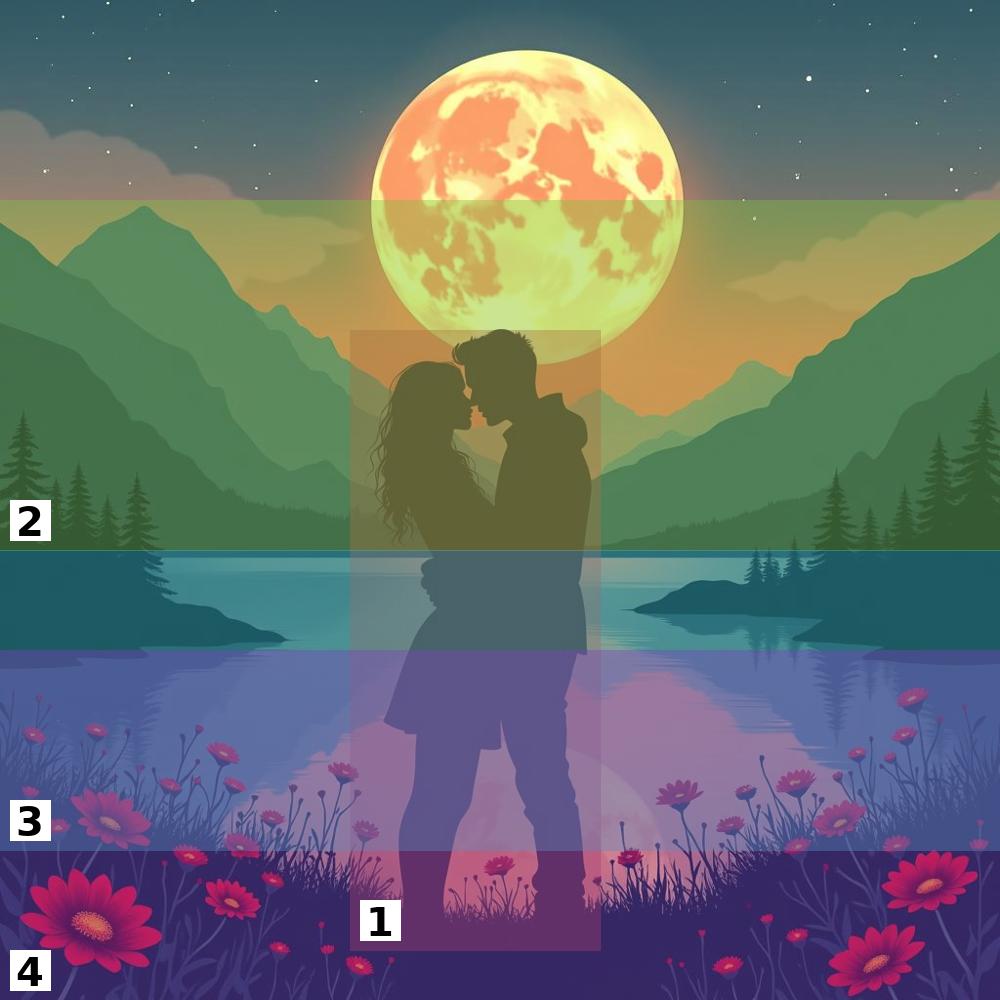} \\

        (d) Gemini-exp-1206 & (e) Claude-3.5-sonnet & (f) Qwen2-VL-72B \\

    \end{tabular}
    \caption{Comparison of image understanding and refining capabilities of multimodal LLMs. Bounding boxes indicate areas requiring modification, with related prompts provided in the Appendix \ref{app:mllm}.}
    \label{fig:Comparison_Bunding_Box}
\end{figure*}

In the interactive algorithms, the understanding model is implemented using a multimodal LLM. For the selection of this model, we conduct a comparative analysis. We test multiple multimodal LLMs using the prompts mentioned in Appendix \ref{app:prompt}. The models include Qwen2-VL-7B \cite{wang2024qwen2}, LLaMa3.2-vision-90b-instruct \cite{dubey2024llama}, InternVL2-26B \cite{chen2024internvl}, Gemini-exp-1206 \cite{team2023gemini}, Claude-3.5-sonnet \cite{TheC3}, and Qwen2-VL-72B \cite{wang2024qwen2}. The locations that each model indicates needed modification are shown in Figure \ref{fig:Comparison_Bunding_Box}, with the corresponding modification prompts described as follows:


\textbf{Qwen2-VL-7B}:
\begin{enumerate}
    \item The couple is beautifully illuminated by the warm glow of the full moon, casting a soft light on their silhouettes.
    \item The mountains in the background are adorned with a gradient of colors, transitioning from deep blues at the base to warm oranges near the peaks, creating a stunning contrast against the night sky.
    \item The serene lake reflects the moon's light, creating a mirror-like surface that enhances the tranquility of the scene.
    \item The trees along the shore are elegantly lit, their dark outlines highlighted by the moonlight, adding depth and dimension to the landscape.
    \item The red flowers surrounding the couple add vibrant pops of color, contrasting beautifully with the cool tones of the night sky and enhancing the overall romantic atmosphere of the scene.
\end{enumerate}

\textbf{LLaMa3.2-vision-90b-instruct}:
\begin{enumerate}
    \item The couple's embrace is surrounded by a halo of soft, golden light that illuminates their faces and casts a warm glow on the surrounding landscape. The light is gentle and romantic, highlighting the love and affection between the two people.
    \item The full moon in the background is a deep shade of orange, with a subtle gradient effect that gives it a sense of depth and dimensionality. The moon's surface is textured with craters and mountains, adding to its natural beauty and realism.
    \item The mountains in the background are covered in a thick layer of mist, which adds to their mystery and allure. The mist is depicted in shades of blue and purple, with touches of pink and orange where it meets the moonlight. The mountains themselves are rugged and rocky, with steep peaks and valleys that add to their dramatic beauty.
    \item The lake in the foreground is calm and peaceful, reflecting the beauty of the surrounding landscape like a mirror. The water is depicted in shades of blue and green, with touches of purple and pink where it meets the moonlight. The lake's surface is dotted with ripples and waves, adding to its sense of movement and life.
    \item The red flowers in the foreground are vibrant and colorful, adding a pop of color to the otherwise muted palette of the image. The flowers are depicted in various stages of bloom, with some fully open and others still budding. They are arranged in a random pattern, adding to their natural beauty and spontaneity.
\end{enumerate}

\textbf{InternVL2-26B}:
\begin{enumerate}
    \item A luminous full moon with a detailed cratered surface, radiating a warm glow, casting a golden light across the scene. The stars twinkle softly in the night sky.
    \item Mountains with jagged peaks, their dark outlines contrasting against the twilight sky, casting deep blue shadows that blend into the cool night.
    \item A serene lake reflecting the full moon, mirroring the silhouettes of the mountains and adding a touch of tranquility with its still waters.
    \item A field of vibrant red flowers, their petals rich and lush, with leaves that add green highlights, creating a striking contrast to the dark silhouettes of the couple.
    \item A dense cluster of coniferous trees, their dark, spiky branches reaching upwards, standing still in the night breeze.
    \item The couple's outlines are refined with fluid lines, their forms bathed in the moonlight, emphasizing the romantic embrace.
\end{enumerate}

\textbf{Gemini-exp-1206}:
\begin{enumerate}
    \item The full moon is a radiant celestial orb, its surface a tapestry of soft, glowing craters and valleys. A gentle, ethereal light emanates from it, casting a silvery sheen over the entire scene. The moon's glow is subtly diffused by a thin veil of high-altitude clouds, creating a halo effect that adds a mystical quality to the night sky. The light reflects off the clouds, painting them with hues of soft orange and dusky pink, creating a breathtaking backdrop for the romantic scene below.
    \item The tranquil lake is a mirror of serenity, its surface as smooth as glass, reflecting the moonlit sky and the surrounding landscape with perfect clarity. The water is a deep, rich blue, tinged with hints of emerald green near the edges. Gentle ripples, caused by a soft breeze, create a mesmerizing dance of light and shadow on the water's surface. The reflection of the full moon on the lake is not just a mere image but a shimmering pathway of light that seems to lead into another world, adding a touch of magic and wonder to the scene.
    \item The couple stands in a tender embrace, their silhouettes outlined against the luminous backdrop. The woman's dress is a flowing, elegant garment made of a soft, shimmering fabric that catches the moonlight, making her appear almost ethereal. Delicate lace details are visible on the sleeves and hem, adding a touch of romantic sophistication. The man is dressed in a finely tailored suit, the fabric a deep, rich color that contrasts beautifully with the woman's dress. The suit's subtle sheen hints at luxury and refinement, and the way it drapes over his form suggests both strength and tenderness. Their pose is intimate and loving, a perfect representation of a romantic moment frozen in time.
    \item The field of red flowers is a vibrant tapestry of color and life. Each flower is a perfect specimen, with velvety petals in a deep, passionate red. Delicate dewdrops cling to the petals, catching the moonlight and sparkling like tiny diamonds. The flowers sway gently in the breeze, creating a sense of movement and vitality. Among the red flowers, there are interspersed tiny wildflowers in shades of white and yellow, adding a touch of delicate beauty and contrast to the scene. The fragrance of the flowers is subtly hinted at, a sweet, intoxicating scent that adds another layer of sensory richness to the romantic atmosphere.
    \item The mountains rise majestically in the background, their peaks shrouded in a soft, ethereal mist that glows faintly in the moonlight. The slopes are covered in a lush forest of pine trees, their dark green needles providing a striking contrast to the lighter hues of the sky and the lake. The trees are rendered in exquisite detail, with each branch and needle clearly defined, creating a sense of depth and texture. The play of light and shadow on the mountains is intricate and captivating, with the moonlight highlighting certain areas while others remain in deep shadow, adding a sense of mystery and grandeur to the landscape.
\end{enumerate}

\textbf{Claude-3.5-sonnet}:
\begin{enumerate}
    \item The night sky is a masterpiece of celestial beauty. Thousands of twinkling stars dot the deep blue-green expanse, creating a mesmerizing tapestry of light. Wispy, ethereal clouds float gracefully across the horizon, their edges tinged with a soft, warm glow from the moon's radiance. The stars seem to dance and shimmer, their light pulsating gently in the vast cosmic canvas, inviting the viewer to lose themselves in the infinite wonders of the universe.
    \item The full moon dominates the scene, a breathtaking orb of celestial magnificence. Its surface is a intricate map of light and shadow, with every crater and maria visible in stunning detail. The moon glows with an intense golden-orange hue, casting a warm, romantic light across the landscape. Its radiance is so powerful that it seems to pulse with an otherworldly energy, bathing everything below in its enchanting luminescence. The moon's reflection on the lake below creates a perfect, unbroken circle of light, as if two cosmic entities are reaching out to touch one another across the expanse of sky and water.
    \item The mountain range is a majestic spectacle of nature's grandeur. Each peak is intricately detailed, with rugged cliffs, deep crevices, and snow-capped summits that catch the moonlight and glisten like diamonds. The silhouettes of the mountains create a dramatic interplay of light and shadow, with each ridge and valley clearly defined. The gradual transition of colors from the deep blue-green of the closer mountains to the misty purple of the distant peaks creates a sense of depth and mystery, inviting the eye to explore every contour and crag of this awe-inspiring landscape.
    \item The serene lake is a mirror of tranquility, its surface a flawless reflection of the sky above. The water is so still that every detail of the surrounding landscape is perfectly replicated, creating a symmetrical masterpiece. Gentle ripples near the shore catch the moonlight, transforming into liquid gold. The lake's edge is adorned with smooth, dark stones that seem to glow with an inner light, their surfaces slick with the kiss of water. The perfect reflection creates an illusion of an infinite world, where sky and water merge into one continuous realm of beauty.
    \item The field of red flowers is a vibrant explosion of color and life. Each blossom is exquisitely detailed, with velvety petals that catch the moonlight and seem to glow from within. The flowers sway gently in a soft breeze, creating a hypnotic dance of shadow and light. Delicate stems and leaves are visible among the blooms, adding depth and texture to the floral carpet. Dew drops cling to the petals, sparkling like tiny diamonds in the moonlight. The rich, deep red of the flowers contrasts beautifully with the cool tones of the surrounding landscape, creating a visual focal point that draws the eye and warms the heart.
    \item The silhouette of the embracing couple is a poignant symbol of love and connection. Their forms are perfectly balanced, creating a harmonious shape against the backdrop of the moon. The outline of their bodies is crisp and clear, with every detail of their posture and the gentle tilt of their heads towards each other visible. The way they hold each other speaks volumes about their intimacy and affection. Their silhouette creates a powerful contrast against the bright background, making them appear as if they are the sole inhabitants of this magical world, lost in their own private moment of romance and wonder.
\end{enumerate}

\textbf{Qwen2-VL-72B}:
\begin{enumerate}
    \item The couple is dressed in elegant, flowing garments that shimmer under the moonlight. The man's suit is tailored to perfection, with intricate embroidery on the lapels, while the woman's dress is adorned with delicate lace and sequins that catch the light. Their embrace is tender and intimate, creating a sense of warmth and connection.
    \item The mountains in the background are majestic and grand, with snow-capped peaks that glisten in the moonlight. The slopes are covered in lush, green forests that add depth and texture to the scene. The sky above is a canvas of stars, with the full moon casting a soft, ethereal glow over the landscape.
    \item The serene lake reflects the full moon and the surrounding mountains, creating a mirror-like surface that adds tranquility to the scene. The water is calm and still, with gentle ripples that dance in the moonlight. The reflections of the moon and the mountains create a mesmerizing effect, enhancing the romantic atmosphere.
    \item The red flowers surrounding the couple are vibrant and lush, with petals that seem to glow in the moonlight. They are arranged in a natural, wildflower style, adding a touch of whimsy and romance to the scene. The flowers are interspersed with delicate greenery and small, twinkling lights that mimic the stars in the sky.
\end{enumerate}


This task not only requires the model to understand image content but also demands precise visual grounding abilities. Consequently, we observed that most multimodal LLMs are unable to provide viable modification suggestions. Among these models, only Claude-3.5-sonnet and Qwen2-VL-72B meet our requirements. Given that Qwen2-VL-72B is an open-source model, we ultimately decided to use it to construct our understanding module.

\end{document}